\documentclass[10pt,journal,compsoc]{IEEEtran}
\ifCLASSOPTIONcompsoc
  \usepackage[nocompress]{cite}
\else
  \usepackage{cite}
\fi
\ifCLASSINFOpdf
\else
\fi
\usepackage[utf8]{inputenc} % allow utf-8 input
\usepackage[T1]{fontenc}    % use 8-bit T1 fonts
\usepackage{hyperref}       % hyperlinks
\usepackage{url}            % simple URL typesetting
\usepackage{booktabs}       % professional-quality tables
\usepackage{amsfonts}       % blackboard math symbols
\usepackage{nicefrac}       % compact symbols for 1/2, etc.
\usepackage{microtype}      % microtypography
\usepackage{xcolor}         % colors
\usepackage{times}
\usepackage{epsfig}
\usepackage{graphicx}
\usepackage{amsmath}
\usepackage{booktabs}
\usepackage{multirow}
\usepackage{bm}
\usepackage{amsfonts}
\usepackage{caption}
\usepackage{subcaption}
\usepackage{verbatim}
\usepackage{wrapfig}

\usepackage{caption}
\usepackage{subcaption}

\newcommand\Vtextvisiblespace[1][.3em]{%
  \mbox{\kern.06em\vrule height.3ex}%
  \vbox{\hrule width#1}%
  \hbox{\vrule height.3ex}}

\begin{document}
\title{RigLSTM: Recurrent Independent Grid LSTM for Generalizable Sequence Learning}
% Ziyu Wang
% , Wenhao Jiang, Zixuan Zhang, Wei Tang, Junchi Yan, Wei Liu, 

\author{Ziyu~Wang,
Wenhao~Jiang,
        Zixuan~Zhang,
        Wei~Tang,
        Junchi~Yan
        % Wei~Liu
\IEEEcompsocitemizethanks{
\IEEEcompsocthanksitem Z. Wang is with Department of Electrical and Computer Engineering, National University of Singapore, Singapore. \protect\\
W. Jiang  is with Guangming Laboratory, Shenzheng, China.\protect\\
Z.~Zhang is with Department of Computer Science, University of Illinois Urbana-Champaign, USA.\protect\\
W.~Tang is with the Department of Statistics, University of British Columbia, UK.\protect\\
J.~Yan is with the Department of Computer Science and Engineering, MoE Key Lab of Artificial Intelligence, 
Shanghai Jiao Tong University, Shanghai, China.\protect\\
\IEEEcompsocthanksitem  Work was done when Z.~Wang worked at Tencent. 
\IEEEcompsocthanksitem The first two authors contribute equally. 
\IEEEcompsocthanksitem W.~Jiang (cswhjiang@gmail.com) is the corresponding author.}% <-this % stops an unwanted space
\thanks{Manuscript received January 19, 2023; revised August 26, 2023.}}


\markboth{Journal of \LaTeX\ Class Files,~Vol.~14, No.~8, August~2015}%
{Shell \MakeLowercase{\textit{et al.}}: Bare Demo of IEEEtran.cls for Computer Society Journals}
\IEEEtitleabstractindextext{%
\begin{abstract}
Sequential processes in real-world often carry a combination of simple subsystems that interact with each other in certain forms. Learning such a modular structure can often improve the robustness against environmental changes. In this paper, we propose recurrent independent Grid LSTM (RigLSTM), composed of a group of independent LSTM cells that cooperate with each other, for exploiting the underlying modular structure of the target task. Our model adopts cell selection, input feature selection, hidden state selection, and soft state updating to achieve a better generalization ability on the basis of the recent Grid LSTM for the tasks where some factors differ between training and evaluation. Specifically, at each time step, only a fraction of cells are activated, and the activated cells select relevant inputs and cells to communicate with. At the end of one time step, the hidden states of the activated cells are updated by considering the relevance between the inputs and the hidden states from the last and current time steps. Extensive experiments on diversified sequential modeling tasks are conducted to show the superior generalization ability when there exist changes in the testing environment.
 Source code is available at \url{https://github.com/ziyuwwang/rig-lstm}.
\end{abstract}

\begin{IEEEkeywords}
Recurrent models, language models, LSTM, sequence learning.
\end{IEEEkeywords}}

\maketitle
\thispagestyle{plain}
 \pagestyle{plain} 

\IEEEdisplaynontitleabstractindextext

% \fi
\IEEEpeerreviewmaketitle

\IEEEraisesectionheading{\section{Introduction}\label{sec:introduction}}

\IEEEPARstart{R}{ecurrent} neural networks (RNNs)~\cite{elman1990finding} have been widely used in research areas need learning time-evolving sequential patterns and characterizing real-world dynamic processes, such as language modeling~\cite{mikolov2010recurrent}, image/video captioning~\cite{vinyals2015show,xu2015show,venugopalan2015sequence,krishna2017dense}, event sequence prediction~\cite{mei2017neural,xiao2017wasserstein}, video generation~\cite{levine2018video,denton2018prior}, and reinforcement learning for intelligent agents~\cite{peter2015drqn,ivan2015attention}.
However, traditional RNNs, such as GRUs~\cite{chung2014empirical} and LSTMs~\cite{hochreiter1997long}, do not explicitly consider the \textit{modular structures} in these tasks, which are very common in real-world applications. 
A typical example could be predictions of ball motions in the game of billiards. Once, a player hits the cue ball, each ball will move independently according to physical laws,  occasionally colliding with other balls or the edge of the table. Thus, each ball can be seen as an independent component, having its own state and only interacts sparsely with others. The state prediction of the whole game at a certain time is a composition of predictions of all these individual components.
Models that consider subsystems of the target task may better capture the underlying compositional structure, and thus better generalize across situations where a small subset of factors change while most of them remain invariant.

To model  modular structures, a possible way is to introduce sparsely interacting independent components into the model. In such model, the components communicate with others occasionally and update their states only when they are involved. A certain component is corresponding to a certain part of the environment. Thus, when the environment changes, only a fraction of relevant components has to be changed. Therefore, models adopt such designs will be robust to the changing environment.

% appropreate inductive biases into the model. It has been

% A model that is capable of uncovering the modular structure and exploiting the underlying regularities would be robust to the changing enviroments.

% changing enviroment

% {\color{red} train test gap, ood}

One of the most intuitive ways to realize the above concern in the temporal sequence related problems is to introduce multiple components into a single recurrent unit. For example,  Relational Memory Core (RMC)~\cite{rmc} exploits a set of memories that can interact via multi-head attention to enhance the ability of complex sequence modeling. 
Recurrent Independent Mechanism (RIM)~\cite{goyal2019recurrent} takes a further step, and employs a set of recurrent cells in one unit. These cells have independent transition dynamics, and will be activated only when they are relevant to the current inputs. The activated cells communicate with the other cells after state updating. In RIM, the inputs to all the activated cell are the same. However, since the cells are independent and they capture different aspects of the target task, it is necessary to provide an efficient way for each cell to select relevant information from the input. Moreover, the activated cells obtain the information of states from other cells via its own hidden state which can absorb relevant information from other cell with communication mechanism. Thus, if a certain cell is not activated in the last time step, it will not know the states of other cells during its own in-cell transition updating. 
Therefore, the information propagation of states among cells is not efficient enough. Lastly, to prevent the problem of vanishing or exploding gradients, RIM exploits residual connections as in \cite{rmc}, which are not related to the input at the current time step and other cells. However, it might be better to employ information from input and other cells to determine how much information should flow from previous hidden state to the current hidden state.

 \begin{figure}[t]
\centering
\includegraphics[width=0.48\textwidth]{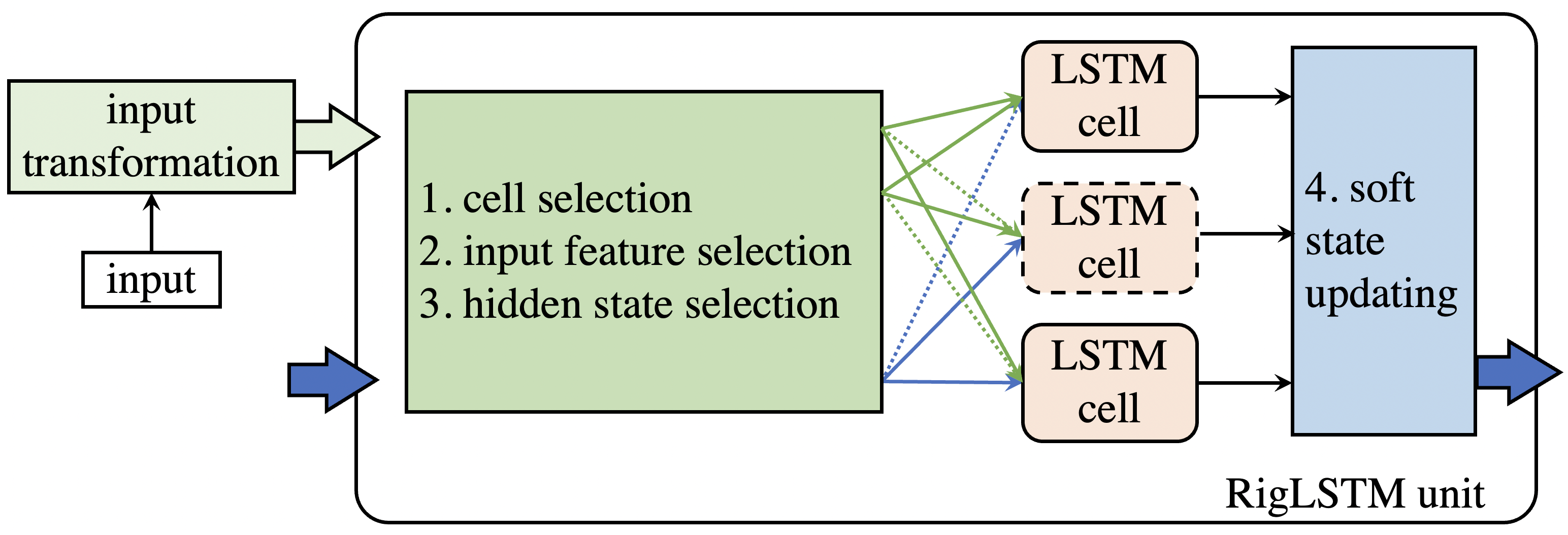}
\caption{An illustration of the unit of recurrent independent Grid LSTM (RigLSTM). The RigLSTM contains multiple LSTM cells, and four steps are performed at each time step inside the unit, \textit{i.e.}, cell selection, input feature selection, hidden state selection and soft state updating. }
\label{fig:rig_lstm_unit}
    % \vspace{-4mm}
\end{figure}

% They communicate occasionally, and will be updated only when they are relevant to the current inputs. {\color{red} The inputs to the related cells are just weighted sum of all inputs.
%  Thus, all inputs are fed into all the RNN cells. However, there might  exist inputs that are not relevant to a certain cell, which will lead to a performance drop. The communication in RIM is performed among only the hidden states, without considering the transition dynamics of the individual cells. Moreover, RIM does not exploit the relationship between hidden states from the previous and current steps to enhance the specialization.
% }

In this paper, based on the above observations on RIM, a novel recurrent unit called recurrent independent Grid LSTM (RigLSTM) is proposed, which is illustrated in Fig.~\ref{fig:rig_lstm_unit}, aiming at facilitating uncovering and exploiting the modular structures for sequence modeling tasks with environment changes during testing.
In the RigLSTM unit, we propose \textit{cell selection}, \textit{input feature selection}, \textit{hidden state selection}, and \textit{soft state updating} to overcome the concerns mentioned above.  The RigLSTM  unit contains multiple LSTM cells and each cell are expected to perform independently. At a certain time step, the input are first transformed into multiple feature vectors, which can be seen as features from multiple views. Different view can provide different information needed by the following steps. Then, the relevance scores between the input features and the hidden states are computed, based on which the cell selection step is performed. Only the activated cells can select relevant input features and hidden states of other cell as input to update their states independently. At last, the soft state updating is employed to facilitate information flow from previous time step to the current time step. 
% the recurrent unit contains a group of LSTM cells. And we propose \textit{input selection}, \textit{hidden state selection}, \textit{cell selection}, and \textit{soft state updating} to overcome the concerns mentioned above. The input selection forces the LSTM cells to choose only relevant input information. The hidden state selection promotes the LSTM cells to select only necessary cells to communicate with, and the communication is performed inside the LSTM cells. At last, the soft state updating, which is a function of all inputs and all cells, is employed to facilitate information flow from previous time step to the current time step. 
Experiments are conducted on multiple sequence modeling tasks across different domains to show that our model has a stronger generalization ability when there exist changes between training and testing environments. The main contributions of the paper are summarized as follows:

\begin{itemize}
\item For generalizable sequence learning, we propose \textit{ input feature selection, cell selection,  and hidden state selection}, integrated with Grid LSTM, to allow for modularization, and decomposition of the input sequence into different subsystems. The specialization and modularization factorize the model into a few simpler and meaningful elements, leading to a more robust system than a homogeneous system~\cite{schmidhuber2018one} and a more light-weighted model than a direct ensemble~\cite{zhang2020diversified}.

\item To ease training for the setting with environment changes during testing, we proposed  soft state updating to propagate information from previous time step. The proposed method considers the current inputs and the states of others cells, which is ignored in other methods.

%and the decoupled subsystems with sparse interaction to each other are more easier to model and generalize.

% 2) To control the incurred additional cost for our enhancement to LSTM, we adopt an inner-product based selection and concatenation module without learnable parameters. Our activation and selection mechanisms are both based on hard decision. These are in contrast to the soft attention in~\cite{goyal2019recurrent}. 

%The overall model size of our model can be controlled, as attention mechanism is devised to allow for specialization on a simple subsystem at one time. 
    %As a result, our new network can decompose the input into different representations that capture different parts of the input and track them through a number of dynamical systems. 
\item Extensive experiments are conducted on diversified tasks and domains to evaluate the promising generalization ability of our approach, which can serve as an enhanced building block for LSTM, at the cost of moderately prolonged inference time. The source code will be made publicly available at the link: \url{https://github.com/ziyuwwang/rig-lstm}.
\end{itemize}

The rest of the paper is organized as follows. Section~\ref{sec:rel} introduces the related work and our main approach is presented in Section~\ref{sec:method}. The extensive empirical studies across different tasks are given in Section~\ref{sec:exp} and Section~\ref{sec:con} concludes this paper.

\section{Related Work}\label{sec:rel}
Our work is related to the modularity of neural network, multiple components and information flow through time in the recurrent models. We will give a brief summarization of these related areas and discuss the relationships with our work.

\subsection{Modularity in Neural Networks}

A network can be designed to be composed of several modules, and each module perform a distinct function. Thus, the whole network can be seen as a combination of module with different specialisms.

In \cite{andreas2016neural}, the authors designed neural module networks for the visual question answering. Several types of module are proposed, and
 questions are decomposed into linguistic substructures, which are then used to dynamically instantiate the network.
A routing mechanism was proposed in \cite{rosenbaum2017routing}  for the setting of multi-task learning. A routing network consists of two components, a router and a set of function blocks. Given an input, the router selects a function block to apply and pass the output back to the router recursively. The whole network is trained with reinforcement learning. Thus, the routing network composes different function blocks for each input dynamically.
Similarly, a modular network was proposed in \cite{kirsch2018modular}. The modular networks model the probability of the modular to be chosen, and the whole network is trained with generalized Viterbi EM, without any artificial regularization. 
In \cite{bahdanau2018systematic}, the relationships between systematic generalization and the module layout were studied on a synthetic VQA dataset. And it was found that end-to-end training do not facilitate systematic generalization and explicit regularization are required.

Most of the above models do not consider multiple activation and relevant input selection at a certain time step. However, these are the considered in our model.

% \cite{shazeer2017outrageously}
% \cite{chang2018automatically}

\subsection{Recurrent Models with Multiple Components}
For complex sequence modeling, many works focus on splitting the memory cell or the hidden state into sub-groups. In Clockwork RNN~\cite{koutnik2014clockwork}, the authors proposed to divide the hidden state into  groups with equal size, which are then updated with different frequencies. The interactions between groups occur periodically. While IndRNN~\cite{li2018independently} treats each single element in the hidden state separately, and the elements are updated completely independently. Grid LSTM~\cite{gridlstm} uses multiple recurrent cells and arranges them in a multidimensional grid. Except for vectors and sequences, it can be even applied to higher dimensional data like images. To enlarge the capacity of recurrent memory, Neural Turing Machines (NTM)~\cite{ntm}, Differential Neural Computers (DNC)~\cite{dnc}, and the more recent Memory Networks (MN)~\cite{initmemnet, memnet} consider introducing external memory modules, which supports reading and writing operations based on the attention mechanism. For an explicit separation of memories, the Relational Memory Core (RMC)~\cite{rmc} introduces a matrix version of memory that is able to incorporate new inputs through a self-attention mechanism. Recurrent Entity Networks (EntNet)~\cite{entnet} develops a dynamic version of external memory for characterizing a sequence in a non-parallel way.

Recently, Recurrent Independent Mechanisms (RIM)~\cite{goyal2019recurrent} employs multiple RNN cells in a single recurrent unit, and it adopts a competing strategy to activate the relevant cells and a communication mechanism to allow cell to know the states of other cells at each certain time step. The designs are also adopted and improved by following works. 
Bidirectional recurrent independent mechanisms (BRIM) \cite{brim}  combines  bottom-up and top-down signals  dynamically using attention, which leads to reliable performance improvements. In \cite{madan2021fast}, the authors combined meta-learning with modular structure, and  showed that meta-learning the modular structure among tasks helps achieve faster adaptation in reinforcement learning.

In RIM, the inactivated cells are updated with default dynamics. The activated cells also communicate with each other to exchange useful information.  However, the irrelevant inputs are also fed into the cells, which leads to bad effects in forming specializations. Moreover, the communication in RIM does not consider the transition dynamic of the individual cells. Our work will exploit cell selection,
input feature selection, hidden state selection, soft state updating that integrated with Grid LSTM to overcome these drawbacks of RIM.

\subsection{Recurrent Models with Information Flow through Time}
To facilitate the training and improving the generalization ability, information from previous hidden states is exploited at the current time step in some specially designed models.
Zoneout~\cite{krueger2016zoneout} randomly keeps some hidden units unchanged from a previous time step.  By preserving some part of hidden units, the information is more easily propagated through time. 
The recurrent highway network (RHN)~\cite{zilly2017recurrent} introduces a gating function to combine the previous hidden state and the current hidden state, which helps propagate the gradient flow. Thus, the recurrent depth can be significantly increased in RHN.
RIM~\cite{goyal2019recurrent} and RMC~\cite{rmc} exploit residual connections from previous hidden states to help propagate information through time.

In this paper, we propose a soft state updating mechanism to combine the hidden states from the previous time step and the current step, by considering the contents of the current input and the states of the other cells. Our method aims to propagate gradient information, and enhance specialization among the cells for the tasks  existing environment changes during testing.

\begin{figure*}[t!]
\centering
\includegraphics[width=0.9\textwidth]{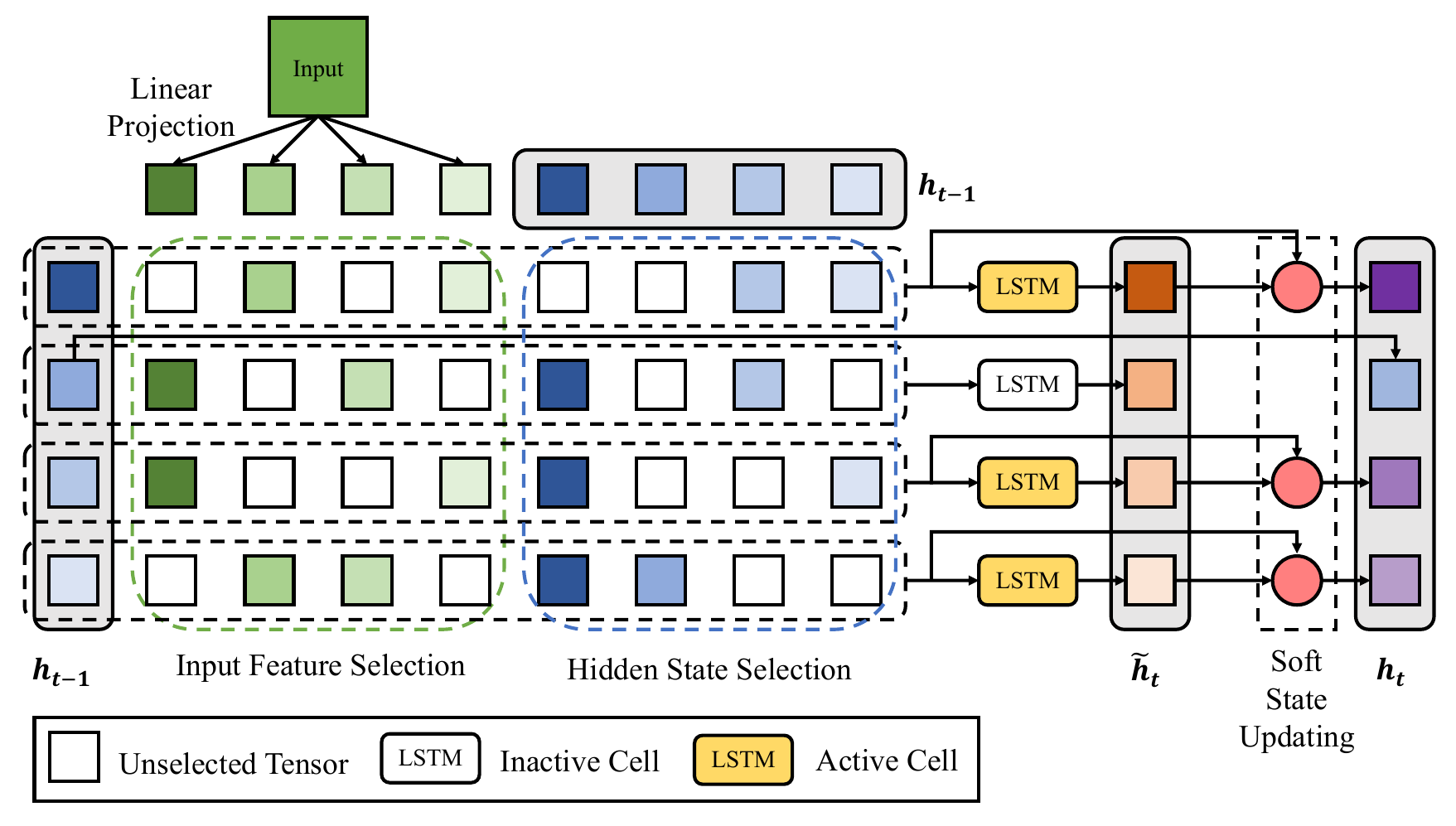}
% \vspace{-5pt}
\caption{\textbf{Recurrent Independent Grid LSTM (RigLSTM).} The input is transformed to multiple tensors and selected by hidden states of each cell. Hidden vectors are selected by similar rules. The selected input and hidden vectors are concatenated and fed into LSTM cells. Only the LSTM cells with the highest sum of similarity scores with input vectors are activated, which are updated by the soft states updating mechanism. It combines newly computed hidden states with old ones.}
\label{fig:main_framet}
%\vspace{-2mm}
\end{figure*}

\section{Recurrent Independent Grid LSTM}\label{sec:method}
In this section, we will give a brief introduction of Grid LSTM~\cite{gridlstm} as background. Then, we will present our model. Lastly, we will discuss the differences between our model and RIM.

\subsection{Preliminaries on Grid LSTM}
For an LSTM cell~\cite{lstm}, the cell comprises a hidden vector $\bm{h}_{t-1} \in \mathbb{R}^{d}$ and a memory vector $\bm{c}_{t-1} \in \mathbb{R}^{d}$, where $d$ is the size of hidden state vector. Taking a vector $\bm{x}_{t} \in \mathbb{R}^{D}$ as input, where $D$ is the dimensionality of the input feature vector, the computation process inside an LSTM cell at time step $t$ is expressed as: 	
% $\bm{f}_{t} = \sigma(\bm{W}_{f} \bm{H}_{t}), \bm{i}_{t} = \sigma(\bm{W}_{i} \bm{H}_{t}), 	\bm{o}_{t} = \sigma(\bm{W}_{o} \bm{H}_{t}), 	\bm{g}_{t} = \operatorname{tanh}(\bm{W}_{g} \bm{H}_{t}), \bm{c}_{t} = \bm{f}_{t} \odot \bm{c}_{t-1} + \bm{i}_{t} \odot \bm{g}_{t}, 	\bm{h}_{t} = \bm{o}_{t} \odot \operatorname{tanh}(\bm{c}_{t})$ 
\begin{align}\label{LSTM}
	\bm{f}_{t} &= \sigma(\bm{W}_{f} \bm{H}_{t}), \\
	\bm{i}_{t} &= \sigma(\bm{W}_{i} \bm{H}_{t}), \\
	\bm{o}_{t} &= \sigma(\bm{W}_{o} \bm{H}_{t}), \\
	\bm{g}_{t} &= \operatorname{tanh}(\bm{W}_{g} \bm{H}_{t}), \\
	\bm{c}_{t} &= \bm{f}_{t} \odot \bm{c}_{t-1} + \bm{i}_{t} \odot \bm{g}_{t}, \\
	\bm{h}_{t} &= \bm{o}_{t} \odot \operatorname{tanh}(\bm{c}_{t}),
\end{align}
where $\sigma$ is the logistic sigmoid function and $\bm{W}_{f}, \bm{W}_{i}, \bm{W}_{o}, \bm{W}_{g} \in \mathbb{R}^{(D + d) \times d}$ are all linear transformation matrices.  $\bm{H}_{t}$ is the concatenation of the input $\bm{x}_{t}$ and the previous hidden vector $\bm{h}_{t-1}$, whihc can expressed as
\begin{align}
    \bm{H}_{t} = \left[\bm{x}_{t}, \bm{h}_{t-1}\right],
\end{align}
% $\bm{H}_{t} = \left[\bm{x}_{t}, \bm{h}_{t-1}\right]$,
where $[\cdot,\cdot]$ is the concatenation operator that stacks column vectors vertically to form a new column vector. The above process is denoted as:
\begin{align}
\bm{h}_{t}, \bm{c}_{t} = \operatorname{LSTM}\left(\bm{H}_{t}, \bm{c}_{t-1}; \bm{W}\right),
\end{align}
% $\bm{h}_{t}, \bm{c}_{t} = \operatorname{LSTM}\left(\bm{H}_{t}, \bm{c}_{t-1}; \bm{W}\right)$, 
where $\bm{W}$ is the set of trainable parameters, \textit{i.e.}, $\bm{W}_{f}, \bm{W}_{i}, \bm{W}_{o}$ and $\bm{W}_{g}$.

Grid LSTM~\cite{gridlstm} is a recurrent unit that contains $N$ LSTM cells,  and each cell update its own states individually. At time step $t$,  we denote the hidden state vectors of the LSTM cells as  $\bm{h}^{1}_{t}, ..., \bm{h}^{N}_{t}$, and memory vectors as $\bm{c}^{1}_{t}, ..., \bm{c}^{N}_{t}$. Given an input $\bm{x}_t$, the Grid LSTM is updated as
\begin{align}
	\bm{h}^{1}_{t}, \bm{c}^{1}_{t} &= \operatorname{LSTM}\left(\bm{H}_{t}, 
	\bm{c}^{1}_{t-1}; \bm{W}^{1}\right), \\ 
	& \ ... , \\
	\bm{h}^{N}_{t}, \bm{c}^{N}_{t} &= \operatorname{LSTM}\left(\bm{H}_{t}, \bm{c}^{N}_{t-1}; \bm{W}^{N}\right),
\end{align}
where $\bm{H}_{t}$ is defined as
\begin{align}
\bm{H}_{t} = \left[\bm{x}_{t}, \bm{h}^{1}_{t-1}, \cdots, \bm{h}^{N}_{t-1}\right].
\end{align}
% $\bm{H}_{t} = \left[\bm{x}_{t}, \bm{h}^{1}_{t-1}, \cdots, \bm{h}^{N}_{t-1}\right]$.
$\bm{W}^{k}$ is a collection of transformation matrices for the $k$-th LSTM cell. Each LSTM cell inside a Grid LSTM has distinct parameter matrices $\bm{W}^{k}_{f}, \bm{W}^{k}_{i}, \bm{W}^{k}_{o}, \bm{W}^{k}_{g}$, and updates with individual dynamics. 
In this way, the Grid LSTM unit enables all internal LSTM cells to communicate with each other while operate independently.

\subsection{Recurrent Independent Grid LSTM}
The overall structure of our model is illustrated in Fig.~\ref{fig:main_framet}. We can see that our model contains several LSTM cells, and each cell is designed to update with independent transition dynamics.  At a certain time step, the input is firstly transformed into multiple vectors to provide different views for the cells, since each cell might need different aspects of the input. Then, a  subset of cells are activated, and each activated cell selects transformed input features and relevant cell states as its inputs to update the state according its own transition dynamics. At last, to facilitate information flow through time, the soft updating is exploited.

The details about our model will be presented in the following subsections. To provide clear descriptions, the notations are summarized in the  Table~\ref{tab:notation}.

\begin{table*}
\centering
    \begin{minipage}{0.80\linewidth}
%   \centering
%   \vspace{-20pt}
\caption{A summery of notations used in the paper.}
	\label{tab:notation}
% \vspace{-5pt}
\resizebox{1\textwidth}{!}{
\begin{tabular}{l|c}
	\toprule
		Notation & Meaning \\
		\midrule
		$d$ & The dimentionality of LSTM hidden states. \\ 
		$N$ & The number of LSTM cells in one unit. \\
        $K$ & The number of input feature vectors. \\
        \midrule
        $K_x$ & The number of selected input feature vectors. \\
        $K_h$ & The number of selected hidden state vectors. \\
        $K_a$ & The number of selected cells. \\
        \midrule
        $S_t^j$ & The set of top $K_x$ inputs. \\
        $R_t^j$ & The set of top $K_h$ hidden states. \\
        $C_t^j$ & The set of top $K_a$ cells selected. \\
        \midrule
        $\mathbf{H}_t^j$ & The input for the $j$-th LSTM cell at time step $t$. \\
        $\mathbf{h}_t^j$ & The hidden state of the $j$-th LSTM cell  at time step $t$ after regular update. \\
        $\tilde{\mathbf{h}}_t^j$ & The hidden state of the $j$-th LSTM cell at time step $t$ after soft updating. \\
        $\mathbf{h}_{t-1}^{k,j}$ & The $k$-th input hidden state for the $j$-th LSTM cell  at time step $t$ after hidden state selection. \\
        $\mathbf{x}_{t}^{k,j}$ & The $k$-th input for the $j$-th LSTM cell at time step $t$ after input selection. \\
        % $\mathbf{h}_t^j$ & \tabincell{c}{The hidden state of the $j$-th LSTM cell \\ at time step $t$ after regular update.} \\
        % $\tilde{\mathbf{h}}_t^j$ &\tabincell{c}{The hidden state of the $j$-th LSTM cell \\ at time step $t$ after soft updating.} \\
        % $\mathbf{h}_{t-1}^{k,j}$ & \tabincell{c}{The $k$-th input hidden state for the $j$-th LSTM cell \\ at time step $t$ after hidden state selection.} \\
        % % $\mathbf{x}_{t}^{k,j}$ & The $k$-th innput for the $j$-th LSTM cell at time step $t$ after input selection. \\
        % $\mathbf{x}_{t}^{k,j}$ & \tabincell{c}{The $k$-th input for the $j$-th LSTM cell \\ at time step $t$ after input selection.} \\
% 		\midrule
    \bottomrule
	\end{tabular}
	}
    \end{minipage}
    %\vspace{-5pt}
\end{table*}

\subsubsection{Input Transformation}
In our model, the cells are expected to be responsible for a certain aspect of the whole system. Thus, the inputs needed for each cell might be different. To facilitate each cell  to form specialization, the input feature is transformed into multiple feature vectors with linear transformation to provide different views of the input for the cells. Ideally, each cell only absorbs information from the views needed. Specifically, the input $\bm{x}_t$ at time step $t$ is first transformed into $K$ vectors $\bm{x}^{1}_{t}, \bm{x}^{2}_{t}, ..., \bm{x}^{K}_{t} \in \mathbb{R}^{d}$, which can be seen as multiple views of the input.

\subsubsection{Cell Selection}
A similarity score between the $k$-th input feature vector and the $j$-th  LSTM cell is computed as: 
% $s_{k,j} =  \bm{x}^{k}_{t} \cdot \bm{h}^{j}_{t},$
\begin{align}\label{x_score}
s_{k,j} =  \bm{x}^{k}_{t} \cdot \bm{h}^{j}_{t},
\end{align}
where $\cdot$ is the inner product between two vectors. 

The proposed model is designed to dynamically select the LSTM cells that are relevant to the current input. For the $j$-th LSTM cell, the relevance score is computed as
\begin{align}
s_j = \sum_k s_{k,j},
\end{align}
where $s_{k,j}$ is the score defined in Eq.~\eqref{x_score}. Thus, the relevance score is the sum of the scores between the $K$ inputs and the LSTM cell state. At each time step, the top $K_{a}$ cells are selected (activated), denoted by $C_t$. 
The hidden states of the selected LSTM cells are updated with the soft state updating, which will be described below. The hidden state vectors and memory vectors of the inactivated LSTM cells remain unchanged.

\subsubsection{Input Feature Selection}
% As shown in Fig.~\ref{fig:input_selection},
At each time step, for the $j$-th LSTM cell, we select the top $K_{x}$ out of $K$ input features according to the similarity scores, and denote the collection of top $K_{x}$ inputs as set $S_{t}^j$. Then, the input feature vectors for $j$-th LSTM cell are determined by the rules:
\begin{equation}
	\bm{x}^{k,j}_{t} = \begin{cases}
	    \bm{x}^{k}_{t}&  \bm{x}^{k}_{t}\in S_{t}^j \\
	    \bm{0}& \bm{x}^{k}_{t}\notin S_{t}^j.
	    \end{cases} 
\end{equation}
The proposed input selection strategy is designed to enhance the diversity of each LSTM cell's input. While the operation of replacing the unselected vectors with all-zeros is for more consistent modeling and easier optimization.

% \begin{figure}[tb!]
% \centering
% \includegraphics[width=0.3\textwidth]{figures/input_selection.pdf}
% % \vspace{-10pt}
% \caption{Input feature vector selection in RigLSTM. For each LSTM cell, input vectors with the top $K_x$ similarity scores (out of $K$) are selected. The unselected are set to zero.}
% \label{fig:input_selection}
% % \vspace{-10pt}
% \end{figure}

% \begin{figure}[tb!]
% \centering
% \includegraphics[width=0.25\textwidth]{figures/soft_states_update.pdf}
% %  \vspace{-10pt}
% \caption{ The soft state updating mechanism devised in this paper. For the $j$-th LSTM cell, all relevant inputs and selected hidden states, except its own hidden state, are concatenated into one vector. The weights for $\tilde{\bm{h}}^{j}_{t-1}$, $\bm{h}^{j}_{t-1}$ are computed by similarity. The final hidden state $\bm{h}^{j}_{t}$ is the weighted sum of $\tilde{\bm{h}}^{j}_{t-1}$ and $\bm{h}^{j}_{t-1}$.}
% \label{fig:soft_update}
% %  \vspace{-10pt}
% \end{figure}

\subsubsection{Hidden State Selection}
Similar to the input feature vector selection, each LSTM cell also selects the hidden states of relevant cells as inputs, to obtain the information from related cells. The selection is  based on the similarity between cells. Since the hidden state vectors have the same dimensionalities, we simply use the inner product between hidden states. For the $j$-th cell, its own hidden vector is to be selected. For the rest $(N-1)$ hidden states, we select the top $K_{h}$, denoted as $R_{t}^j$. For the $j$-th LSTM cell, we define
\begin{equation}
	\bm{h}^{k,j}_{t-1} = \begin{cases}
	\bm{h}^{k}_{t-1 }& \bm{h}^{k}_{t-1 }\in R_{t}^j \ \text{or} \ k = j \\
	\bm{0} & \bm{h}^{k}_{t-1 } \notin R_{t}^j
	\end{cases} 
\end{equation}
and by combining with the input feature selection, we obtain the input vector $\bm{H}^{j}_{t}$ for the $j$-th LSTM cell:
\begin{equation}
	\bm{H}^{j}_{t} = \left[\bm{x}^{1,j}_{t}, \cdots, \bm{x}^{K,j}_{t}, \bm{h}^{1,j}_{t-1}, \cdots, \bm{h}^{N,j}_{t-1} \right].
\end{equation}

\subsubsection{Soft State Updating}

\begin{figure}[t!]
\centering
\includegraphics[width=0.45\textwidth]{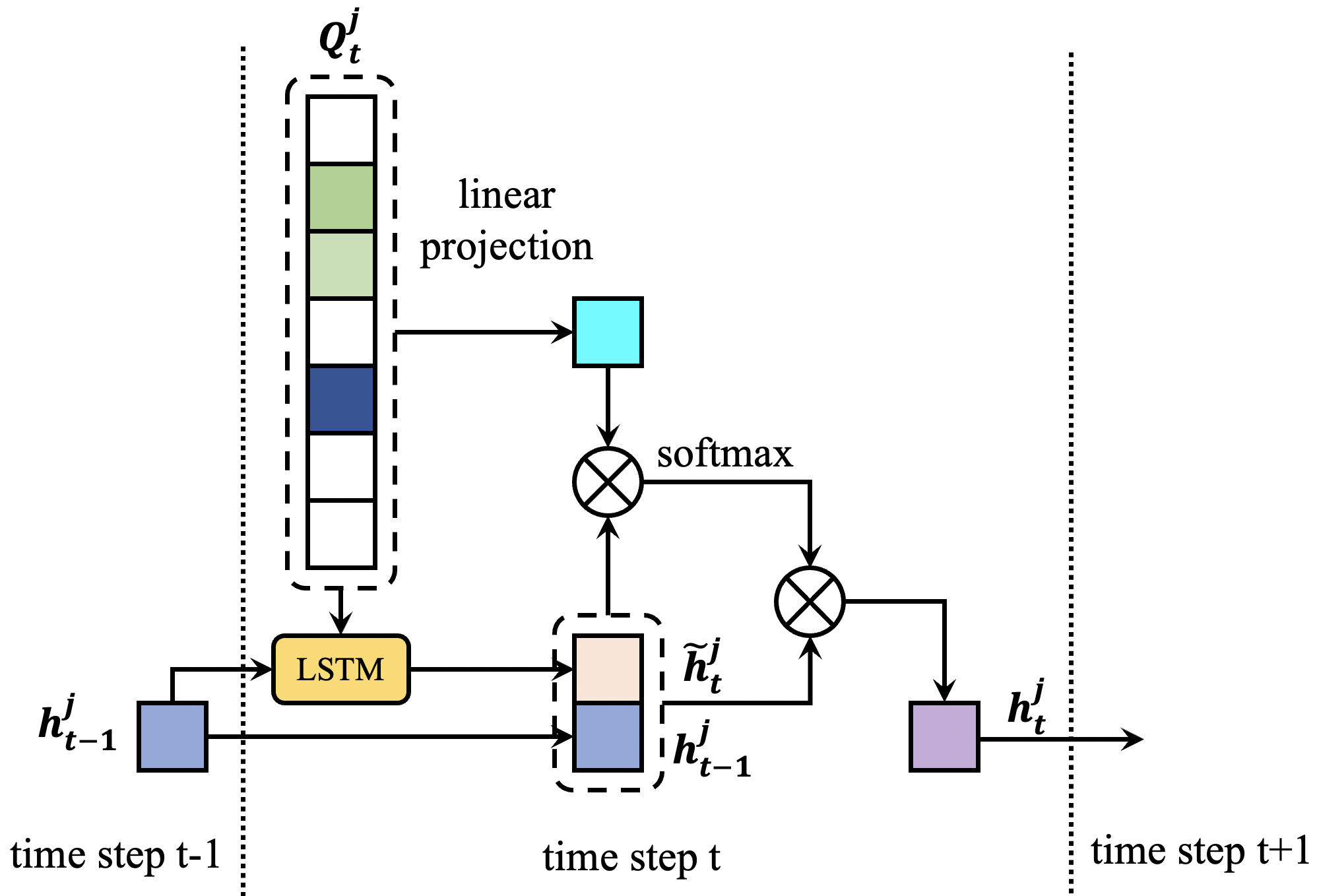}
% \vspace{-10pt}
\caption{An illustration of the soft state updating mechanism proposed in this paper. For the $j$-th LSTM cell, all relevant inputs and selected hidden states, except its own hidden state, are concatenated into one vector.  Then the weights for $\tilde{\bm{h}}^{j}_{t}$ and $\bm{h}^{j}_{t-1}$ are computed by considering the similarities between them. The final hidden state $\bm{h}^{j}_{t}$ is a weighted sum of $\tilde{\bm{h}}^{j}_{t}$ and $\bm{h}^{j}_{t-1}$.
}
\label{fig:soft_update}
\end{figure}

To facilitate information propagation through time and to ease the training, we propose a soft state updating mechanism for activated cells, which is illustrated in Fig.~\ref{fig:soft_update}.
% The soft state updating aims to combine the previous hidden states and the current hidden states. 

For the $j$-th LSTM cell, after the regular hidden state update step, we can obtain the hidden states $\tilde{\bm{h}}_{t}$, which is expressed as:
\begin{align}
    \tilde{\bm{h}}^{j}_{t}, \bm{c}^{j}_{t} &= \operatorname{LSTM} \left(\bm{H}^j_{t}, \bm{c}^{j}_{t-1}; \bm{W}^{j}\right).
\end{align}

All the selected vectors of the $j$-th cell except for $\bm{h}^{j,j}_{t-1}$ are concatenated to form $\bm{Q}^{j}_{t}$, as given by:
\begin{align}
 \bm{Q}^{j}_{t} = \left[\bm{x}^{1,j}_{t}, \cdots, \bm{x}^{K,j}_{t}, \bm{h}^{1,j}_{t-1}, \cdots, \bm{h}^{j-1,j}_{t-1}, \bm{h}^{j+1,j}_{t-1}, \cdots, \bm{h}^{N,j}_{t-1}\right]. 
\end{align}

Then, we stack the hidden states from the previous step and the current step to form a matrix
% $\bm{K}^{j}_{t} = \left[ \bm{h}^{j}_{t-1};\tilde{\bm{h}}^{j}_{t} \right]$,
\begin{align}
\bm{K}^{j}_{t} = \left[ \bm{h}^{j}_{t-1};\tilde{\bm{h}}^{j}_{t} \right],
\end{align}
where $[\cdot; \cdot]$ is an operator for $n$ column vectors of size $d$ to form a $d \times n $ matrix. We express the final hidden state as 
% $\bm{h}^{j}_{t} = \bm{K}^{j}_{t}\operatorname{softmax}\left(  \left(\bm{K}^{j}_{t}\right)^{\top}  \bm{W}_q \bm{Q}^{j}_{t} \right),$
\begin{align}
    \bm{h}^{j}_{t} = \bm{K}^{j}_{t}\operatorname{softmax}\left(  \left(\bm{K}^{j}_{t}\right)^{\top}  \bm{W}_q \bm{Q}^{j}_{t} \right),
\end{align}
where $\bm{W}_q$ is linear transformation matrix. We can see that the combination weights for the previous states and current state are determined by the input feature vectors and  the hidden state from all the cells. Thus, the information propagation through time in our model is effected by the context information, which will be showed to be effective for the setting of generalizable sequence learning.

\subsection{Differences with RIM}

% Compared with RIM, except the soft state updating method which is an extra part, the other three parts of our model can be regarded as targeted improvements to the corresponding structure of RIM. We discuss the difference from RIM in the following three ways:

RIM contains three steps, \textit{i.e.}, cell selection, independent hidden state updating and communication among hidden states.  The relevant cells are activated by computing the similarity between the hidden states and the input. Only the cells of high similarity with the current input are activated. The activated cells update their state with independent transition dynamics. At last, the updated cells communicate with all the other cells via attention mechanism. We will discuss the differences between our model and RIM on the following aspects in detail.

\subsubsection{On Inputs}
The input of each LSTM cell in RIM is a weighted sum of an original input and a null input, implying that inputs of the LSTM cells are the same vector multiplied by different coefficients. Thus, the information fed into cells are the same. However, our RigLSTM removes the null input, and transforms the original input into multiple features vectors, each of which captures different information of the original input. Then, each cell selects relevant feature vectors by similarity between the transformed feature vectors and the corresponding hidden states. Therefore, the information fed into the cells in our model are all different. Such distinctness in input will facilitate the cells to form specialization.

% With this setting and input selection, the inputs of LSTM cells are different combinations of vectors obtained by projection, resulting stronger diversity of inputs as well as facilitating the specialization of cells.

\subsubsection{On Retrieving Context Information From Other Cells}
% \subsubsection{On State Updating and Communication Steps}
% 对于来自其他cell的context information处理不同。rim 先更新状态，然后利用attention机制来为
% state 补上 context information。更新状态的时候其实没有主动获取context information。比如：。。。。
% 我们的方法在更新状态的时候有context information。因此输出的state直接含有这个信息，不需要额外的 communication。比如。。。。
% 我们方法将状态更新和cell 间通讯合并成一个步骤，可以更有效利用context information。但是RIM 是分离的，利用效率低下。

\begin{figure*}[tb!]
    \centering
    \begin{subfigure}[b]{0.49\textwidth}
    \centering
    \includegraphics[width=1\textwidth]{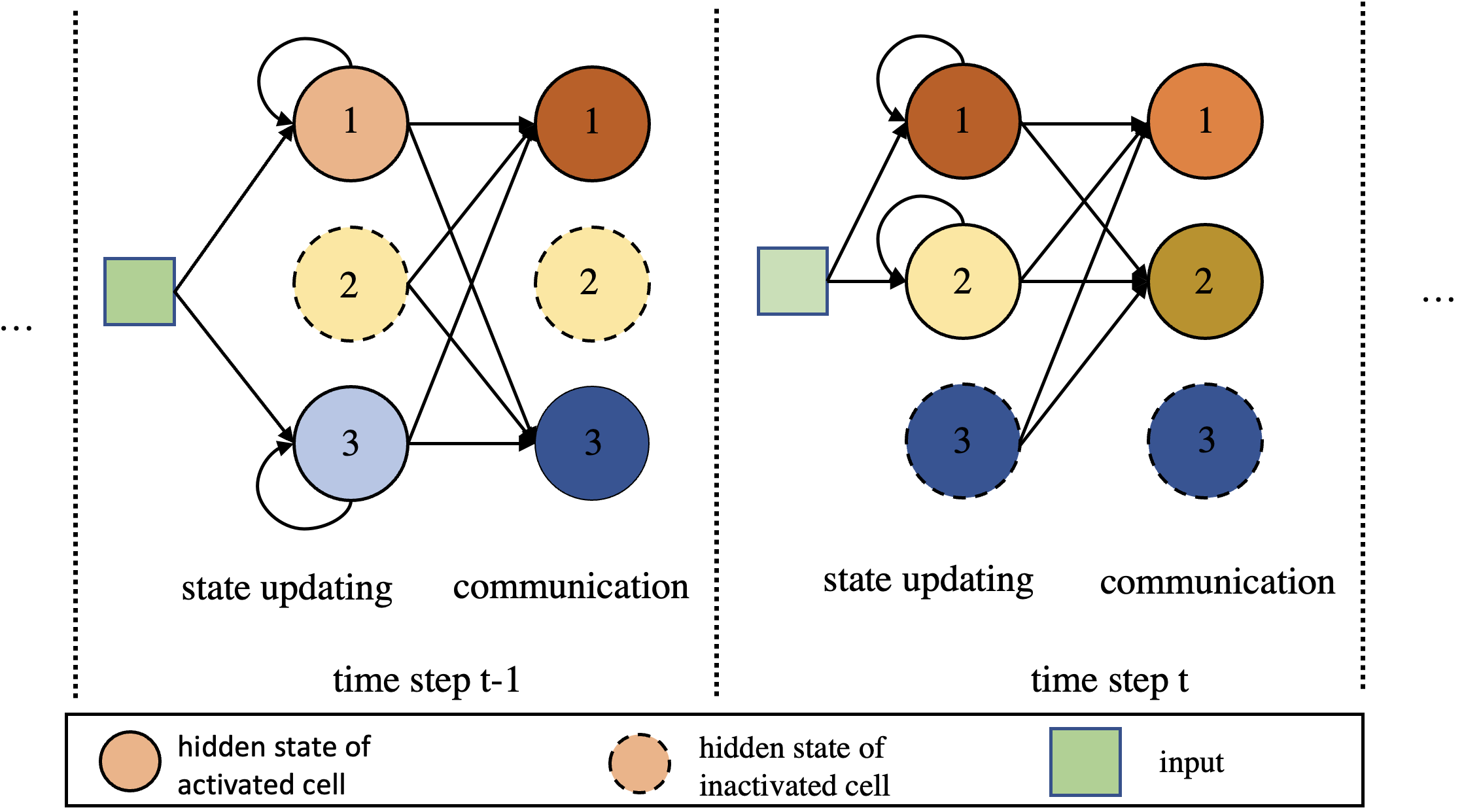}
    \caption{An illustration of retrieving context information in the state updating in RIM.
    Cell 1 is activated at time step $t$ and $t-1$, and cell 2 is only activated at time step $t$. When performing state updating at time step $t$,  cell 1 knows the state of other cells from the communication step performed at time step $t-1$. Cell 2 does not have the chance since it is not activate at time step $t-1$. Thus, the state updating step of cell 2 lacks updated context information.  Thus, each cell might have outdated context information. }
    \label{fig:diff_rim}
     \end{subfigure}
     \hfill
      \begin{subfigure}[b]{0.49\textwidth}
    \centering
    \includegraphics[width=0.83\textwidth]{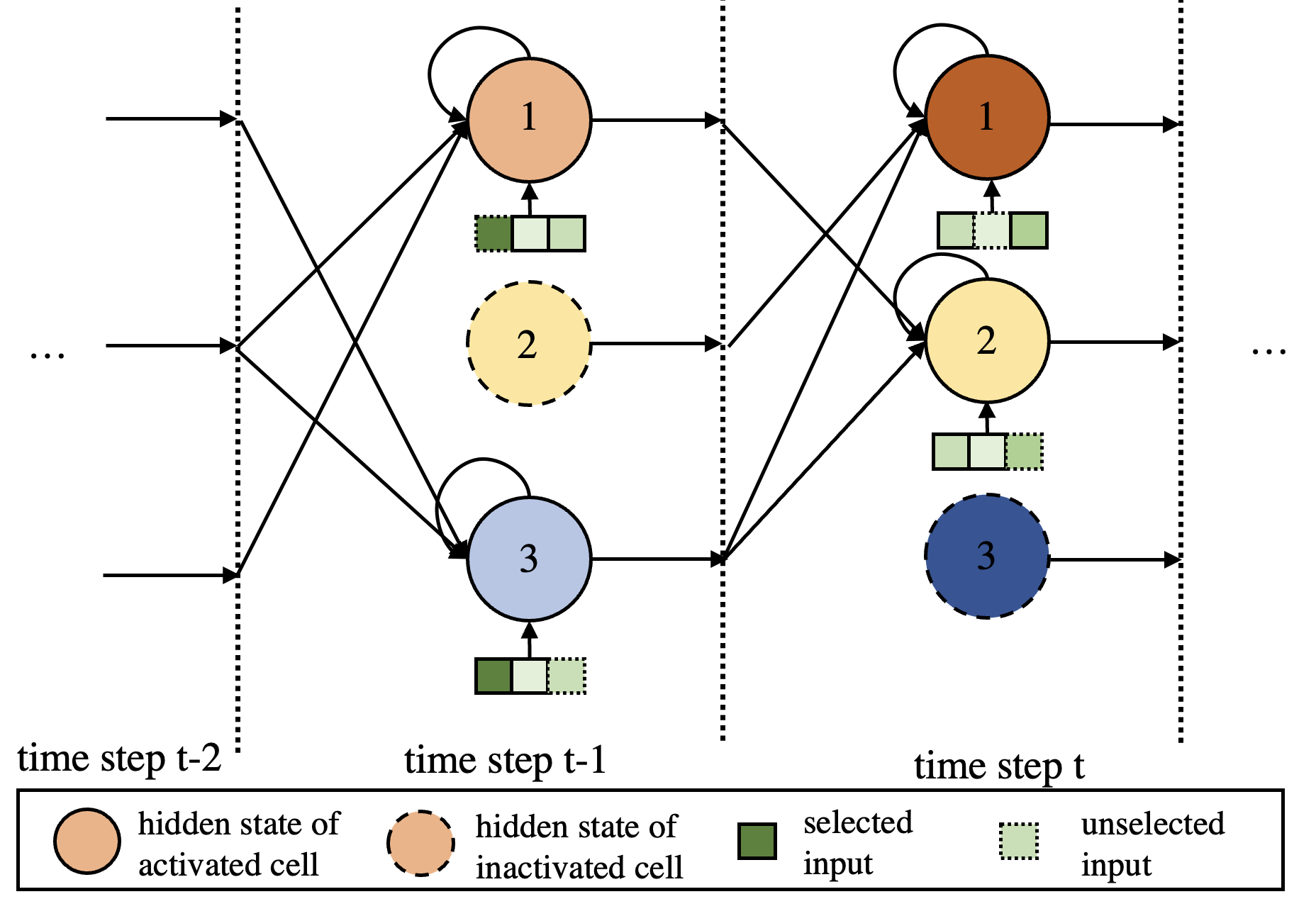}
    %\vspace{-15pt}
    \caption{An illustration of retrieve context information during state updating in the proposed RigLSTM. Each cell selects relevant hidden state of the other cells as input to perform state updating. Even if cell 2 is not activated at time step $t-1$, it still knows the updated state information from all the other cells. Thus, each cell has updated context information. }
    \label{fig:diff_rig_lstm}
     \end{subfigure}   
    \caption{Comparisons of the ways how to retrieving context information from other cells when performing state updating steps in between RIM and RigLSTM. }
    \label{fig:diff}
    % \vspace{-2mm}
\end{figure*}

% The ways that how cells retrieve necessary information from other cells in the state updating step in RIM and our model are also different. The differences are illustrated in Fig~\ref{fig:diff}.
The cells in the recurrent unit need to know the states of other cells, so as to cooperate with each other.
In RIM, each activated cell performs communication step after updating their own state independently.
The cells do not know the states of other cells when performing state updating. 
However, the state updating step and communication step are combined into one single step in our proposed unit.  And the cells in our model update their states knowing the context information.
% The differences are illustrated in Fig~\ref{fig:diff}.

In RIM, a cell which is activated at the current time step but not activated at the previous time step will lack updated context information when performing state updating. An example is shown in Fig.~\ref{fig:diff_rim}. Cell-1 is activated at time step $t$ and $t-1$. Therefore, cell-1 have the context information from other cells implicitly at time step $t$, since the hidden state of cell-1 can retrieve necessary information via communication at time step $t-1$, which will be fed into cell-1 at time step $t$. However, cell-2 does not have a chance to retrieve such information. Cell-2 is only activated at time step $t$, and no retrieving (communication) operation is performed at time step $t-1$. The state of cell-2 is updated with outdated context information.

% Thus, actions depending on the hidden state of cell 2 will not be that robust for lacking updated context information from other cells.

In RigLSTM, each activated cell selects hidden states from other cell as inputs. Thus, they all know the necessary context information. As shown in Fig.~\ref{fig:diff_rig_lstm}, cell-2 at the time step $t$ knows the states of cell-1 and cell-2. 
Therefore, the actions depending on the hidden state of cell-2 is more reliable.
The effectiveness will be illustrated with experiments  in the following section.

% Unlike using attention to gather input information and hidden states for an LSTM cell in RIM, RigLSTM utilizes vector concatenation for both input vectors and hidden states via the structure of Grid LSTM, forming a larger vector and retaining more information for each LSTM cell. This can be viewed as a stronger communicating and sharing function between LSTM cells against RIM.

% \subsubsection{On Communication Step}
% In RIM, the communication step occurs after the activated cell hidden state updating step. The communication adopted by RIM is a kind of attention~\cite{bahdanau2014neural}, and the hidden states from activated cells query information from all the other cells. The communication step is the only way for a cell to know the states of other cell. 

% In RigLSTM, the 

\subsubsection{On Information propagation through Time}
In RIM, for activated cells, the information of hidden state from previous time step is propagated to the current time step with residual connections. 
% For inactivated cells, the hidden states are kept unchanged. 
However, in our model, the soft state updating mechanism considers all relevant input features and hidden states from other cells to determine the combination coefficients. Thus, the inputs and the states of other cell can effect information flow through time for the activated cells. Such design can benefit the forming of specialization of the cells, which will be illustrated in the following section.

% Hidden states that have little relevance to an LSTM cell are suppressed by small weights of an attention mechanism in RIM. Information selection mechanisms in RigLSTM set the irrelevant information to zeros, which increases the suppression to these irrelevant vectors, and magnifies the difference between cells upon, thus benefiting the specialization of cells.

\section{Experiments}\label{sec:exp}
We conduct experiments on both synthetic and real-world benchmarks to show the advantages of our models, especially the generalization ability when there exist some changes in testing environments. 

% The detailed settings of the experiment environment are given in Appendix.

\subsection{Experimental Settings}

The experiments are conducted on a single machine with 256G memory and 96 CPUs (Intel(R) Xeon(R) Platinum 8255C CPU @ 2.50GHz). The models trained with a single NVIDIA Tesla V100 GPU. Four representative tasks are used to compare the performances of competing models. For all the experiments, we use Adam as the optimizer and set the learning rate to 0.0001, if not otherwise specified.

%, \textit{i.e.}, different image resolutions for sequential image classification, different dormant phase lengths in digit memorization, different numbers of balls for video prediction, and different levels of gaming difficulty for reinforcement learning.

% \begin{table}
% \centering
%     % \begin{minipage}{0.6\linewidth}
% %   \centering
% %   \vspace{-20pt}
% \caption{A summery of hyperparameters used in the experiments.}
% 	\label{tab:mnist_ablation}
% % \vspace{-5pt}
% % \resizebox{1\textwidth}{!}{
% \begin{tabular}{l|c}
% 	\toprule
% 		Parameter & Value \\
% 		\midrule
% 		 Optimizer & Adam~\cite{adam} \\
% learning rate  &  $1\times 10^{-4}$ \\
% batch size for Se  & \\
% cell dimension & \\
% % 		\midrule
%     \bottomrule
% 	\end{tabular}
% % 	}
%     % \end{minipage}
%     %\vspace{-5pt}
% \end{table}

%\vspace{-1mm}
\subsection{Sequential Image Classification Task}
%\vspace{-1mm}
In line with~\cite{gridlstm,goyal2019recurrent}, we consider the task of sending a sequence of pixels of an image into a recurrent model,  and predicting its label. The generalization ability is evaluated on data with resolutions different from the training set. Experiments are designed based on the two well-know dataset: MNIST and CIFAR-10. 
% Since the upsampled images contain more distracting and uninformative regions, a more modular model should have a better ability to ignore these patterns and thus generalize better to larger resolutions.

\subsubsection{Experimental Settings}
For MNIST, the resolution of  training data is set to $14 \times 14$, while the competing models are evaluated at three higher resolution settings ($16 \times 16$, $19 \times 19$, and $24 \times 24$). For  CIFAR-10,  the resolution for training is  $16 \times 16$, and
the   resolutions for testing are $19 \times 19$, $24 \times 24$ and $32 \times 32$. Images of various resolutions are obtained by performing a nearest-neighbour sampling method on original images. The images of MNIST~\cite{lecun1998gradient} are converted into binary digits (0 or 1), and then embedded to dimension of 600. For the colorful images in CIFAR-10~\cite{cifar}, the three channels, of which pixel values are from 0 to 255, are separately embedded to dimension 200 and then concatenated to form the inputs. For RIM, the number of cells in one unit is set to 6. For RigLSTM, we set  $K$ and $N$ to 6, and $K_x = K_h = K_a$. The dimensionalities of single cells are both set to 100. 
For fair comparisons, the LSTM model has a hidden size of 600. We use the same configuration for the other experiments if not otherwise specified.

\begin{table}
\centering
\begin{minipage}{0.98\linewidth}
  \centering
\captionof{table}{
Test accuracy ($\%$) on the sequential MNIST dataset. All models are trained on images with resolution $14 \times 14$, and evaluated at three resolution settings. $N$ is the number of cell in one unit, and $K_a$ is the number of activated cell at a certain time step.
}
% \small
\label{tab:mnist}
% \vspace{-5pt}
\resizebox{1\textwidth}{!}{
    \begin{tabular}{l|c|c|c|c}
		\toprule
		Models  & ($N, K_a$)  & $16 \times 16$ & $19 \times 19$  & $24 \times 24$ \\
		\midrule
		LSTM~\cite{lstm} & - &84.5 &52.2 &21.9 \\ % running
		RIM~\cite{goyal2019recurrent} & (6, 4) & 90.0 & 73.4 & 38.1 \\
		EntNet~\cite{entnet} & - & 89.2 & 52.4 &23.5 \\
		RMC~\cite{rmc} & - & 89.58 & 54.23 &27.75 \\
		Transformers~\cite{transformer} & - & \textbf{91.2} & 51.6 & 22.9 \\
		GridLSTM~\cite{gridlstm}  & - & 88.32 & 38.1 & 20.84 \\%~\cite{gridlstm}
		BRIM~\cite{brim} & - & 88.6 & 74.2 & 51.4 \\
        \midrule
        RigLSTM & (6, 5) & 89.50 & \textbf{80.71} & \textbf{59.60} \\
% 		\specialrule{0em}{1pt}{1pt}
% 		\hline
% 		\specialrule{0em}{1pt}{1pt}
% 		\hline
		\bottomrule
	\end{tabular}
}
\end{minipage}
\end{table}

\subsubsection{Analysis}
% The best results are achieved with the optimal super-parameter $K_{a} = 4$. 
The experimental results of all models on MNIST dataset are presented in Table~\ref{tab:mnist}. We can see that our model greatly improves the performance over the competing methods on most of the settings, especially when the changes on the testing environment are large. 
Note that almost all the models achieve the performance of about 90\% for the $16 \times $16 setting, which is a relative easy setting, because the changes between training and testing are small.
For the other harder settings, our model achieved the best results. And the gaps between our model and all the other models are quite large. 
The reuslts on CIFAR-10 are shown in Table~\ref{tab:cifar}. 
Similar phenonimenon can also be observed as on MNIST dataset. Thus, our model has the best generalization ability for the setting with environment changes.

\begin{figure}[tb!]
    \centering
    \includegraphics[width=0.46\textwidth]{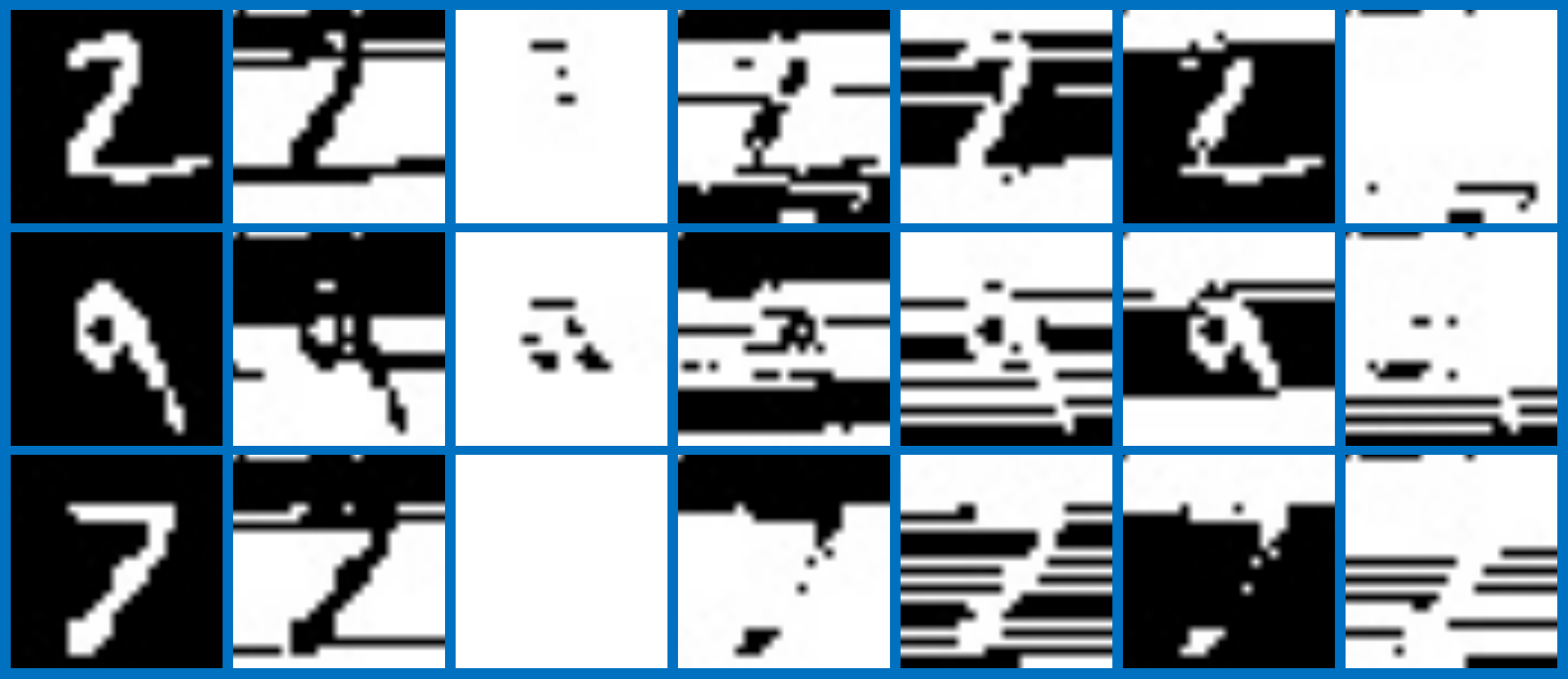}
  %  \vspace{-10pt}
    \caption{Examples of cells activation patterns when testing three digit images with resolution $24 \times 24$ on MNIST dataset. The first column is the digit images, whose pixels are fed into models in the scan line order. Columns $2$ to $7$ show the activation image of 6 LSTM cells, respectively. A white pixel indicates that the LSTM cell is activated at the corresponding time step, while a black pixel indicates not activated.
    % Different cells show different (binary mask) activation patterns.
    }
    \label{fig:cells_example}
\end{figure}

\begin{figure}[tb!]
    \centering
    \includegraphics[width=0.48\textwidth]{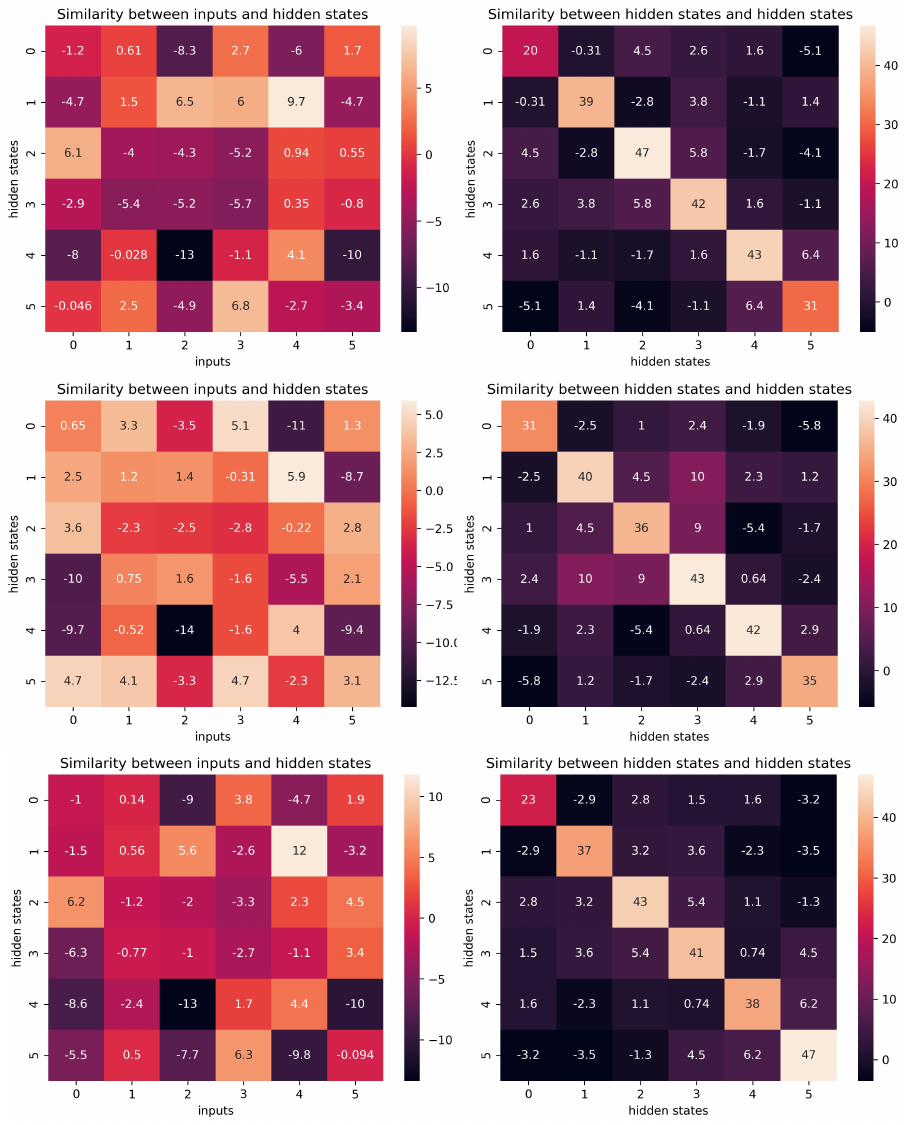}
    %\vspace{-15pt}
    \caption{Examples of input selection scores (left column) and hidden state selection scores (right column). The pairs of scores are from the $289$-th time step of the three cases in Fig.~\ref{fig:cells_example}.}
    \label{fig:hs_example}
\end{figure}

We list three examples of cell selection results during testing in Fig~\ref{fig:cells_example}. And the corresponding input feature selection scores and hidden state selection scores at the $289$-th time step are shown in Fig.~\ref{fig:hs_example}. From Fig~\ref{fig:cells_example}, it can be seen that the $5$-th cell (cell-4) is activated mostly on the pixels corresponding to stroke of digit, while the $1$-st cell (cell-0) is activated in the opposite way. It seems that cell-4 is responsible for capturing digit strokes and  cell-0 is responsible for capturing the background. This confirms our hypothesis that each LSTM cell in RigLSTM is responsible for different functions and the whole network is well modularized. From Fig.~\ref{fig:hs_example}, we can see that the similarity scores are quite diverse. The similarity pattern between inputs and cells is more stable than that between hidden states. For example, cell-1 favours the input-4, and cell-4 dislike input-2. The relationship between cells are more complex. It is natural because the inputs from different views provide quite different information, and different cells might need information from a certain view through all  time steps. Moreover, the states of cells are effected by the inputs. It is not easy to form a stable relationships between cells. Results in Tables~\ref{tab:mnist} and~\ref{tab:cifar} and Figures~\ref{fig:hs_example} and ~\ref{fig:cells_example} indicate that our design helps the cells form specializations, thus our model improve the generalization performance compared with other methods.

% {\color{red} more text about fig~\ref{fig:cells_example} \ref{fig:hs_example}}

% \begin{figure*}[tb!]
%     \centering
%     \begin{subfigure}[b]{0.49\textwidth}
%     \centering
%     \includegraphics[width=0.98\textwidth]{figures/cells_activation_example.pdf}
%     \caption{Examples of cells activation on resolution $24 \times 24$ on MNIST for three digits. The first column is the original inputs, which are fed into LSTM models in a scanning order. Columns $2$ to $7$ show the activation situation of LSTM cells $1$ to $6$, respectively. A white (resp. black) pixel indicates that the LSTM is activated (resp. not activated) at the corresponding time step. Different cells show different (binary mask) activation patterns.}
%     \label{fig:cells_example}
%      \end{subfigure}
%      \hfill
%       \begin{subfigure}[b]{0.49\textwidth}
%     \centering
%     \includegraphics[width=0.98\textwidth]{figures/xh_scores.pdf}
%     %\vspace{-15pt}
%     \caption{Examples of input selection scores (left column) and hidden state selection scores (right column). The pairs of scores are from the $289$-th time step of the three cases in Fig.~\ref{fig:cells_example}.}
%     \label{fig:hs_example}
%      \end{subfigure}   
%     \caption{-----------------------------}
% \end{figure*}

\begin{table}
\centering
    \begin{minipage}{1\linewidth}
   \centering
\caption{Testing accuracy ($\%$) on sequential CIFAR-10 dataset. All models are trained on resolution $16 \times 16$ and evaluated at 3 larger resolutions.  $N$ is the number of cell in one unit, and $K_a$ is the number of activated cell at a certain time step.
% $N$ and $K_a$ denote the number of total and activated cells, respectively.
}
\label{tab:cifar}

\resizebox{1\textwidth}{!}{
\begin{tabular}{l|c|c|c|c}
		\toprule
		Models  & ($N, K_a$)  & $19 \times 19$ & $24 \times 24$  & $32 \times 32$  \\
		\midrule
		LSTM~\cite{lstm} & - & 54.4 & 44.0 & 32.2 \\ % running
		RIM~\cite{goyal2019recurrent} & (6, 4) & 56.9 & 51.4 & 40.1 \\
		RMC~\cite{rmc} & - & 49.9 & 44.3 & 31.3 \\
		Transformers~\cite{transformer} & - & 58.5 & 50.4 & 43.6 \\
		GridLSTM~\cite{transformer} & - & 56.6 & 46.4 & 35.3 \\
		BRIM~\cite{brim} & - & 60.1 & 57.7 & 52.2 \\
        \midrule
        RigLSTM &(6, 4)	& \textbf{60.5} & \textbf{59.5} & \textbf{54.9} \\
% 		\specialrule{0em}{1pt}{1pt}
		\bottomrule
	\end{tabular}
	}
    \end{minipage}
   % \vspace{-5pt}
\end{table}

\subsubsection{Ablation Study}

\begin{table}
\centering
    \begin{minipage}{0.98\linewidth}
%   \centering
%   \vspace{-20pt}
\caption{Ablation studies on sequential MNIST dataset. Top-1 classification accuracy (\%) of variants of RigLSTM  are reported.}
	\label{tab:mnist_ablation}
% \vspace{-5pt}
\resizebox{1\textwidth}{!}{
\begin{tabular}{l|c|c|c}
	\toprule
		Models & $16 \times 16$ & $19 \times 19$ & $24 \times 24$ \\
		\midrule
        {RigLSTM} & 89.50	&\textbf{80.71}	&\textbf{59.60} \\
		\midrule
	    $\text{RigLSTM}_{{no}\_{input}\_{trans}}$ & 86.32 & 55.28 & 35.75 \\ 
	    \midrule
		$\text{RigLSTM}_{all\_cells}$ & \textbf{91.07} &77.13 &54.01 \\  % 33 or  13(6, *, *)
		 $\text{RigLSTM}_{rand\_cells}$ & 84.13 & 73.27 & 50.39 \\  % 33 or  13(6, *, *)
		 \midrule
		$\text{RigLSTM}_{all\_inputs}$ & 85.48 & 71.43 & 50.61 \\ % (*, 6, *)
		$\text{RigLSTM}_{rand\_inputs}$ & 83.5 & 70.26 & 48.84 \\ % (*, 6, *)
	    $\text{RigLSTM}_{soft\_inputs}$ & 86.74 & 73.29 & 53.97 \\
	    \midrule
		$\text{RigLSTM}_{all\_h}$ & 84.73 & 71.2 & 49.71 \\ % (*, *, 6)
		$\text{RigLSTM}_{rand\_h}$ & 85.24 & 73.56 & 48.35 \\ % (*, *, 6)
	    $\text{RigLSTM}_{soft\_h}$ & 87.35 & 74.83 & 51.38 \\
	    \midrule
		$\text{RigLSTM}_{no\_soft\_state\_updating}$ &90.34 & 75.86 & 54.44 \\  % 23
% 		\specialrule{0em}{1pt}{1pt}
% 		\hline
% 		\specialrule{0em}{1pt}{1pt}
% 		\hline
    \bottomrule
	\end{tabular}
	}
    \end{minipage}
    %\vspace{-5pt}
\end{table}

We also conduct ablation studies on the sequential MNIST dataset to analyze the effects of input transformation, input feature selection, hidden state selection, cell selection, and soft state update. We revise our model by changing or removing one component at a time to obtain a new model, and test the performance of the new model.  The input feature selection and hidden state selection method in our model is a kind of hard selection method, \textit{i.e.}, a candidate of input feature or hidden states can only be selected of unselectied. To verify the necessarity, we design variants that select all input feature or hidden state, or soft selection, \textit{i.e.}, weighting the  candidates with normalized scores. The details of these variants are described as follows:
\begin{itemize}
    \item $\text{RigLSTM}_{{no}\_{input}\_{trans}}$: The input transformation step is removed, and the original input is fed into all the cells.
    \item $\text{RigLSTM}_{all\_cells}$: All cells are activated.
    \item $\text{RigLSTM}_{rand\_cells}$: $K_a$ cells are randomly activated.
    \item $\text{RigLSTM}_{all\_inputs}$: All inputs are selected for each LSTM cell.
    \item $\text{RigLSTM}_{rand\_inputs}$: $K_x$ input features are randomly selected for each LSTM.
    \item $\text{RigLSTM}_{soft\_inputs}$: Select all inputs, and each is weighted by the similarity scores normalized with softmax operation.
    \item $\text{RigLSTM}_{all\_h}$: All hidden states are selected for each LSTM.
    \item $\text{RigLSTM}_{rand\_h}$: $K_h$ hidden states are randomly selected for the LSTM cells.
    \item $\text{RigLSTM}_{soft\_h}$: Select all hidden states, and each of them is weighted with similarity scores normalized with softmax operation.
    \item $\text{RigLSTM}_{no\_soft\_state\_updating}$: Remove the soft state updating step. The hidden state at the end of time step $t$ for the $j$-th cell is $\tilde{\bm{h}}^j_t$.
\end{itemize}

 The performances of the above models are shown in Table~\ref{tab:mnist_ablation}. We can see removing or changing any components will achieve worse performance. Thus, all the components, \textit{i.e.}, input transformation, cell selection, input feature selection, hidden state selection and soft state updating are all necessary for the RigLSTM model. 
 The  variants with random selection achieve almost the worse performances. It is natural that since it is not possible for the model to determine which  input feature or cells will be fed into the model at each time step.
 Moreover, soft selection method is better than all selection method. For example, the performances on the three test setting of $\text{RigLSTM}_{soft\_h}$ are better than 
$\text{RigLSTM}_{all\_h}$, and $\text{RigLSTM}_{soft\_inputs}$ is also better than $\text{RigLSTM}_{all\_inputs}$. Thus, selecting relevant information is necessary for the setting existing environment changes when testing. And hard selection is better than soft selection. At last, we can see that our proposed soft state updating mechanism also improve the generalization performance, by comparing $\text{RigLSTM}_{no\_soft\_state\_updating}$ and RigLSTM. The reason might be that the combination of current state and previous states are determined by all the relevant information, including input features, hidden state of other cells. Such design helps the cells to know the context, thus improves the generalization ability.

% The results show that any random selection would lead to a performance degradation, supporting the effectiveness of our similarity-based method. For the soft scores selection method, we just replace the selection scores of 1 or 0 in RigLSTM with the softmax scores along the dimension of the number of LSTM cells. Compared to hard-decision method, selection and aggregation based on soft scores lead to obvious performance degradation. Fig~
% \ref{fig:hs_example}  visualizes the scores for input selection and hidden state selection at the central ($289$-th) time step of the examples in Fig.~\ref{fig:cells_example}. As the input selection scores of each LSTM cell are different, there would be little chance that LSTM cells produce the same representation, which further enhances the diversity of our inputs.

%The input selection step helps RigLSTM to absorb information only from related views. Moreover, our model also selects the related cells to communicate with. Thus, the input information is carefully selected for each cell, which helps the cell to focus on the information that they are specialized in. At last, the output hidden states are enhanced by combing the previous hidden states and the current hidden state. The weights are determined by the relevance scores to the input. The final output hidden states are forced to be related to the selected inputs to some extent. Therefore, RigLSTM facilitates the LSTM cells to absorb certain views of input,and enhances them to output hidden states related to the input.

\subsubsection{Complexity Analysis}
In this subsection, we compare the inference speed, number of parameters, GPU memory occupation of RigLSTM, LSTM and RIM.
For fairness,  the same sizes of hidden state  are used, and the metrics are obtained on the the sequential MNIST dataset. The results are shown in Table~\ref{tab:mnist-time}.
We can see that the number of parameters of our model is a little larger than the others, which is mainly due to the increase of input dimensionality in each LSTM cell.
For GPU memory usage, our model is better than RIM because the input selection and hidden state selection of LSTM cells are performed sequentially in our implementation, not in an parallel way. This leads to the extension of inference time. We implement RigLSTM in a very straightforward way for easy code reading and modifying, and the inference speed can be improved by more efficient implementations. Overall, RigLSTM trades a small increase in complexity for a significant performance improvement.

\begin{table}
\centering
    \begin{minipage}{1\linewidth}
   \centering
%   \vspace{-10pt}
\caption{Inference time (ms), number of parameters and GPU memory occupation of different models tested on the sequential MNIST dataset. The batch size is 64.}
	\label{tab:mnist-time}
% \vspace{-5pt}
\resizebox{1\textwidth}{!}{
\begin{tabular}{l|c|c|c|c|c|c}
		\toprule
		\multirow{2}{*}{{Models}} & \multirow{2}{*}{${(N, K_a)}$} & \multicolumn{3}{|c|}{{{Inference Time}}} & \multirow{2}{*}{{\#Para}} & \multirow{2}{*}{{Memory}} \\ \cline{3-5}
		&  & $19 \times 19$ & $24 \times 24$  & $32 \times 32$ &  &  \\
% 		Models & ($N, K_a$) & $19 \times 19$ & $24 \times 24$  & $32 \times 32$ & \#Para & Mem \\
		\midrule
		LSTM~\cite{lstm} & - & 66.4 & 93.5 & 149.1 & 2.8M & 1621M \\ % running
		RIM~\cite{goyal2019recurrent} & (6, 4) & 661.2 & 945.5 & 1502.4 & 3.0M & 2604M \\
        \midrule
        RigLSTM &(6, 4)	& 1133.2 & 1590.4 & 2550.2 & 3.6M &1915M \\
% 		\specialrule{0em}{1pt}{1pt}
% 		\hline
% 		\specialrule{0em}{1pt}{1pt}
% 		\hline
    \bottomrule
	\end{tabular}}
    \end{minipage}
    %\vspace{-5pt}
\end{table}
%\vspace{-1mm}

\subsection{Bouncing Ball Video Prediction Task}

Similar to previous works~\cite{goyal2019recurrent,balls}, we evaluate our model by a synthetic ``bouncing balls'' task\footnote{\href{https://github.com/sjoerdvansteenkiste/Relational-NEM}{https://github.com/sjoerdvansteenkiste/Relational-NEM}}, in which multiple balls of different sizes and masses move with an initial setting of velocities.
These balls are in motion under the basic laws of Newtonian Dynamics, where each ball will either keep the original movement state or change the state when a collision with the wall or other balls occurs.

\subsubsection{Task Description}
At each time step, the input is a 64$\times$64 image frame that indicates the current states of the balls. And each pixel is a 0-1 integer scalar, where one indicates that the pixel is part of a ball, and otherwise zero.
The task is similar to video prediction, where we aim at predicting the next frame given previous ones. We utilize a CNN-based encoder and decoder for input feature extraction and output frame generation, respectively, in which the details of the encoder and decoder are exactly the same as those used in~\cite{balls}. Finally, the binary cross-entropy (BCE) loss between predicted and ground-truth frames is adopted for optimization and evaluation.

\subsubsection{Experimental Setting}
We adopt LSTM~\cite{lstm}, GridLSTM~\cite{gridlstm}, Recurrent Independent Mechanisms (RIM)~\cite{goyal2019recurrent}, and Recurrent Entity Networks (EntNet)~\cite{entnet} as the baseline models in our experiment. As the authors of RIM did not provide source code for this task, we conduct all the experiments based on our own implementation. Thus, we achieve the best performance of baseline RIM with hyper-parameters different from RIM paper. The dimensionality of input for all recurrent models is 512, and that of hidden state for LSTM is 250. The number of cells for EntNet, RIM, GridLSTM and RigLSTM is 5, with the individual hidden size of 50. There are three datasets of different numbers of balls and settings, abbreviated as \textbf{\emph{4 balls}}, \textbf{\emph{6-8 balls}}, and \textbf{\emph{curtain}}, respectively. In specific, in \emph{4 balls} dataset, there are four balls moving and colliding during the whole process, while in \emph{6-8 balls} dataset the number could vary from 6 to 8. In the \emph{curtain} dataset, a random rectangular curtain is applied to three moving balls in each frame during the time steps, which may occlude some certain balls at specified time steps. Such a dataset with occlusion is utilized to evaluate the transfer or generalization capability of the models. In each dataset, we have 50,000 sequences for training, and 10,000 sequences each for validation and testing, where each sequence contains 51 time steps of frames. We use learning rate \textbf{0.0003} to train each model until convergence (\textit{i.e.}, the validation loss does not decrease in 5 consecutive epochs). For testing, we feed the first 15 frames to generate 10 frames, and evaluate the performance with the loss on these 10 frames.

\subsubsection{Analysis}
\begin{figure}[tb!]
    \centering
        \begin{subfigure}{0.48\textwidth}
        \centering
        \includegraphics[width=\textwidth]{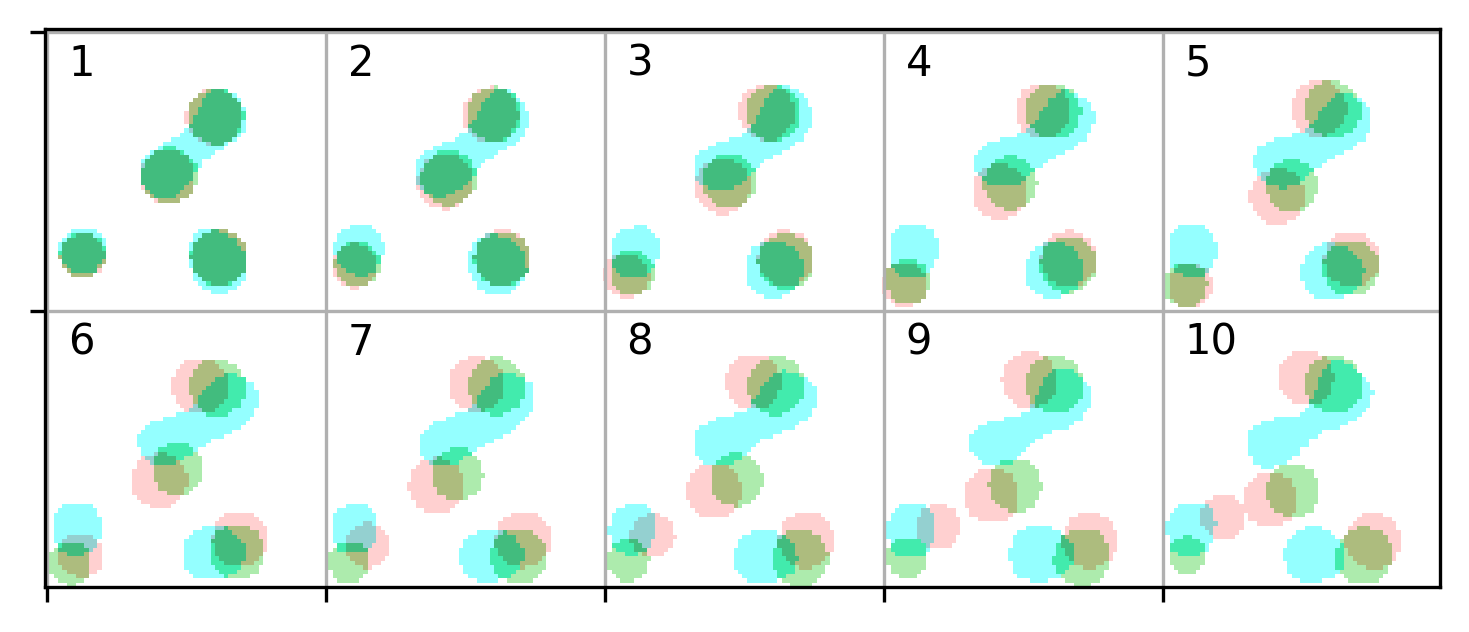}
        \caption{Train on 4 and test on 4.}
    \end{subfigure}
    \begin{subfigure}{0.48\textwidth}
        \centering
        \includegraphics[width=\textwidth]{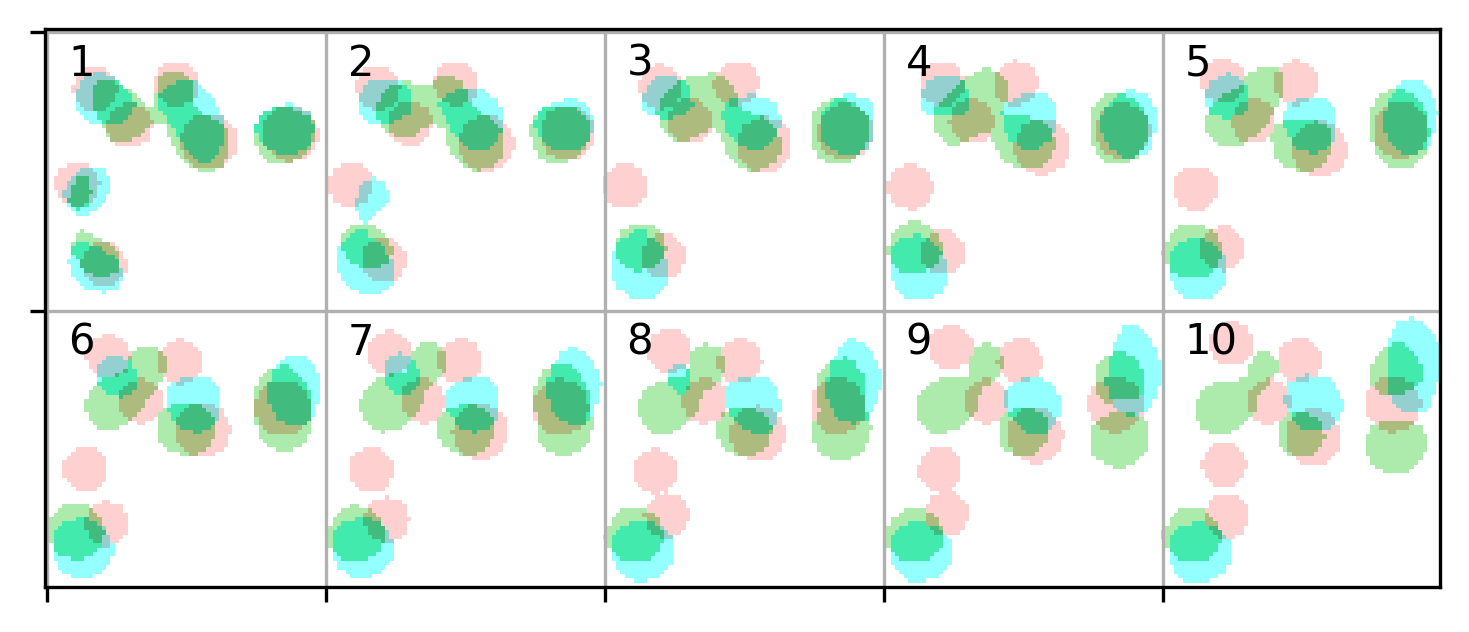}
        \caption{Train on 4 and test on 678.}
    \end{subfigure}
    \begin{subfigure}{0.48\textwidth}
       \centering
        \includegraphics[width=\textwidth]{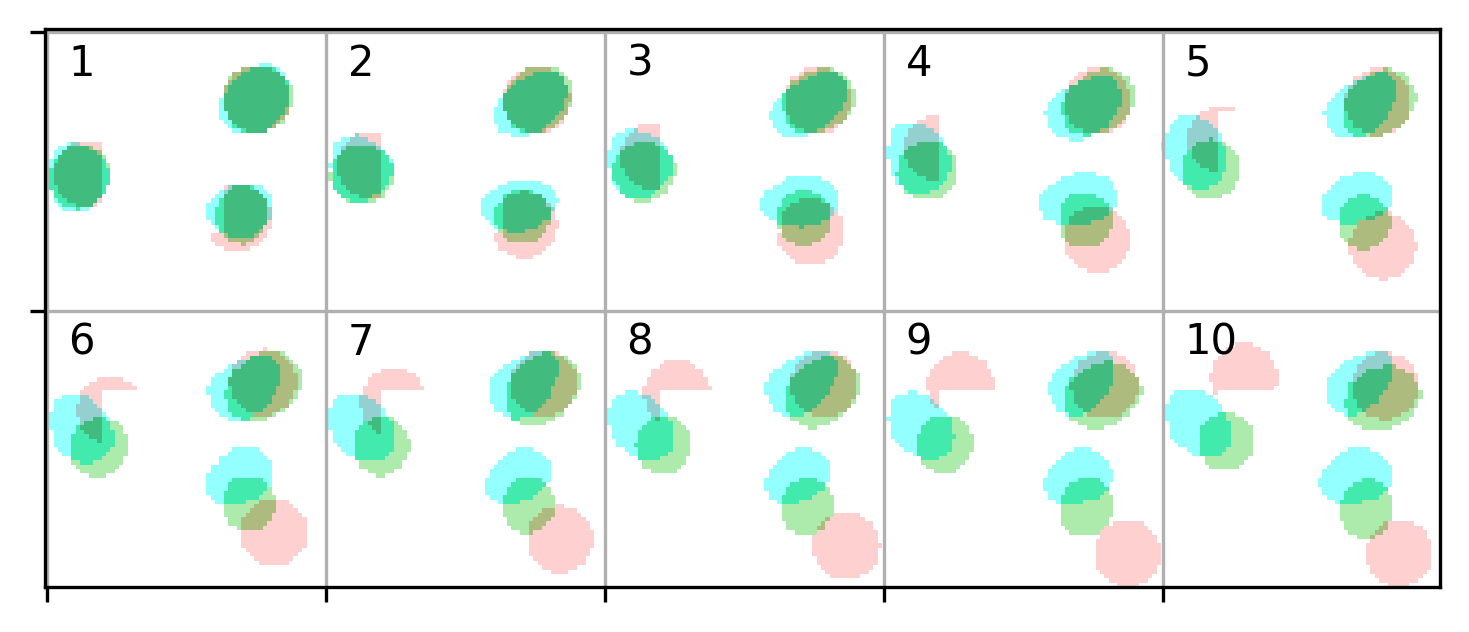}
        \caption{Train on 4 and test with curtain.}
        \label{fig:three sin x}
    \end{subfigure}
    % \vspace{-5pt}
    \caption{Evaluation of RigLSTM and RIM on the Bouncing Ball experiments. The pink, green and blue balls denote predictions of ground-truth, RigLSTM and RIM, respectively.
    }
    \label{fig:ball}
\end{figure}

We report the best test BCE loss in 5 runs on different datasets in Table~\ref{tab:ball-test}: models with multiple cells are usually better than model with only one cell. Thus, employing multiple cells in one unit is necessary for the robustness to the changes in the testing environment. Moreover, our model achieves the best performance in most of the settings, which suggests that the advantages of our novel recurrent unit. To better understand how our model outperforms RIM, we visualize a sample in Fig.~\ref{fig:ball}, which contains 10 frames of ground-truth and predicted results after feeding 15 frames into the models. We compare the predicted frames from RigLSTM (green), RIM (blue) to the ground-truth (pink balls) and see which model gives prediction closer to the ground-truth. As we can see, both RigLSTM's and RIM's results move slower than the ground truth, but RigLSTM's balls can catch up faster comparing to the RIM baseline. Also, the results of RigLSTM are more like balls comparing to some oval-shaped results of RIM. 

%%%%%%%%%%%%%%%%%%%%%%%%%%%%%% ball pics %%%%%%%%%%%%%%%%%%%%

%%%%%%%%%%%%%%%%%%%%%%%%%%%%%% ball pics end %%%%%%%%%%%%%%%%%%%%

\begin{table}
\centering
    \begin{minipage}{0.98\linewidth}
   \centering
    % \vspace{-30pt}
	\centering
	\caption{Roll-out BCE error (\%) of bouncing-balls video prediction. We train each model on \emph{4 balls} and \emph{6-8 balls} datasets, and evaluate both in-distribution and out-of-distribution performances. Bold denotes the best on each task. For RIM~\cite{goyal2019recurrent} and RigLSTM~\cite{gridlstm}, we set $N=5$, $K_a=2$ for the total number of cells and number of activated cells, respectively.}
	\label{tab:ball-test}
% \vspace{-5pt}
\resizebox{.98\textwidth}{!}{
	\begin{tabular}{l|c|c|c|c|c|c}
		\toprule
		Training  & \multicolumn{3}{|c|}{\textbf{{4 balls}}} & \multicolumn{3}{|c}{\textbf{{6-8 balls}}}   \\
% 		\midrule
		\specialrule{0em}{1pt}{1pt}
		\cline{0-6}
		\specialrule{0em}{1pt}{1pt}
		Testing  & {{4 ball}} & {6-8 ball} & {curtain} & {{6-8 ball}} & {4 ball} & {curtain}  \\
		\midrule
        RIM~\cite{goyal2019recurrent}	&32.37	&57.11	&34.81	&32.64	&44.29	&33.71   \\
		EntNet~\cite{entnet}&35.37	&66.63	&36.76	&34.91 &	53.99	&41.76    \\%~\cite{entnet} 
		LSTM~\cite{lstm} &39.23	&72.76	&43.99	&\textbf{28.79}	&54.53 &51.73  \\%~\cite{hochreiter1997long}
		GridLSTM~\cite{gridlstm} &38.36 &72.02 &43.22 &37.36 &58.1 &47.45   \\
        RigLSTM   &\textbf{30.52} &\textbf{49.19} &\textbf{31.72} &31.53 &\textbf{42.64} 	 &\textbf{33.51}   \\
	\bottomrule	
	\end{tabular}}
    \end{minipage}
    % \vspace{-5pt}
\end{table}

\subsection{Reinforcement Learning on MiniGrid}

To evaluate our model on more complex problems, we conduct experiments on reinforcement learning (RL), in which the inputs and the environments are often complicated and challenging.

\subsubsection{Task Description}
Our experiments are conducted on OpenAI Gym-Minigrid\footnote{\href{https://github.com/maximecb/gym-minigrid}{https://github.com/maximecb/gym-minigrid}}, the world of which is an N$\times$M grid of tiles. Each tile can only contain one object, where an agent could take some certain actions at each time step, \textit{e.g.,} turn left, turn right, move forward, pick up, drop off, \textit{etc.}  For example, in a \emph{Door Key} environment, an agent needs to pick up a key and and carries it to open the locked door in order to finish the game. In MiniGrid, there are many other  environments for training an agent to take actions to accomplish such simple tasks. Therefore, it is a suitable platform to evaluate RigLSTM's ability of learning a policy in various environments. In order to use reinforcement learning algorithms to guide the agent, we need first to define a reward for it. Note that given a time step $t$, the current situation is binary, \textit{i.e.}, job finished or not. We can give the agent a reward of one if it accomplishes the task within $F_{max}$ frames, otherwise, the reward is zero. In this way, the averaged reward can also be viewed as the success rate of the task. 

\begin{figure*}[t!]
\centering
\includegraphics[width=0.98\textwidth]{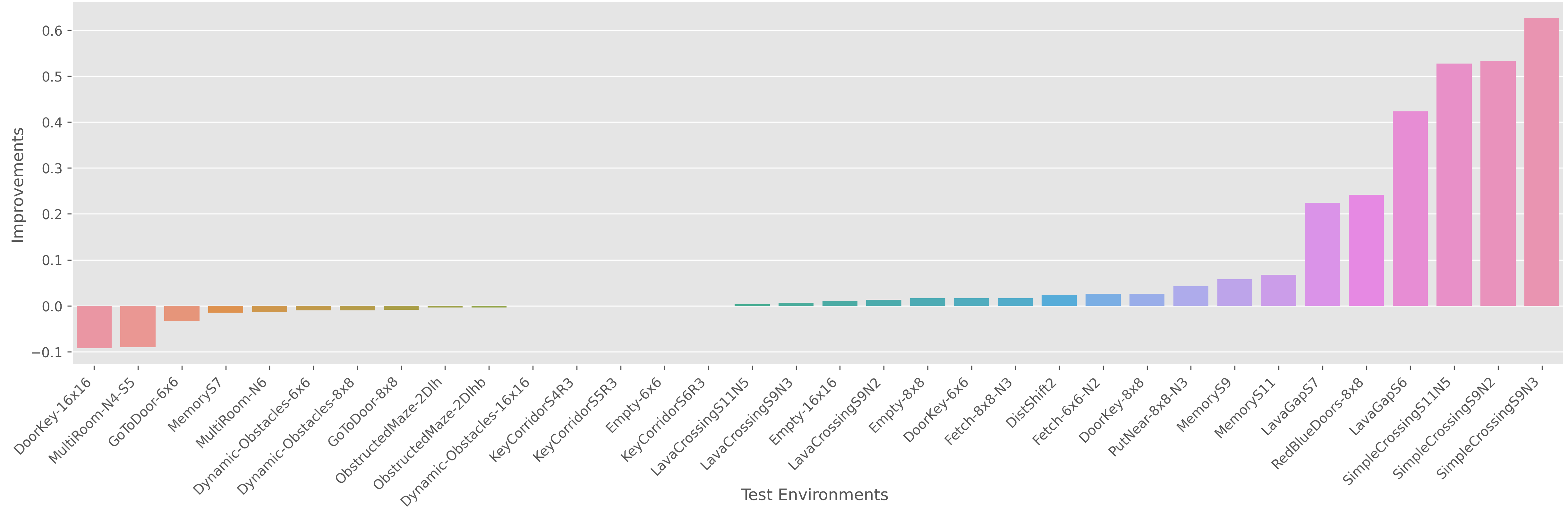}
% \vspace{-5pt}
\caption{Relative value of rewards of RigLSTM-PPO over RIM-PPO on different tasks in MiniGrid. The mean of three trials is reported for fair comparison.
}
\label{fig:rl_reward}
% \vspace{-3mm}
\end{figure*}

\subsubsection{Experimental Settings}
Experiments of this task are also conducted on our own implementation since no source code provided from the authors of RIM. We train our model and RIM~\cite{goyal2019recurrent} across all MiniGrid environments that have different levels of difficulty. All of the models are trained by Proximal Policy Optimization (PPO)~\cite{schulman2017proximal}. To compare the generalization ability, we train them on the easiest environment and test their performance in all difficulty levels of that game, and then compare the mean rewards. For example, \emph{DoorKey} environment has four difficulty levels, $5 \times 5$, $6 \times 6$, $8 \times 8$, $16 \times 16$. We only train the model on \emph{DoorKey-$5 \times 5$} environment and test it on all of the four environments. To alleviate the effect of randomness, we conduct the experiments for each model-environment pair for three times with different random seeds and compare the mean rewards. We compare our RigLSTM based PPO (RigLSTM-PPO) with RIM based PPO (RIM-PPO). We use hyper-parameters ($N=4, K_a=2$) that perform best in most of the experiments conducted by the original RIM implementation for RIM-PPO, and ($N=4, K_a=4, K_x=4, K_h = 4$) for RigLSTM-PPO, while all the other hyperparameters are the same for both models.

\subsubsection{Analysis}
The final results are shown in Fig.~\ref{fig:rl_reward}. The values in the figure are improvements on rewards comparing to that of RIM-PPO. We can see that using our RigLSTM in PPO can significantly increase the rewards the agents can get on most environments. For example, RigLSTM has the highest improvements in the SimpleCrossing environment, where the agent needs to cross many walls to reach the target point. Comparing to other game settings, small SimpleCrossing environment is generalized to larger environments in the way that the agent only needs to notice where the target is and then moves towards it by avoiding walls. In such a situation, the superiority of the input and hidden states selection mechanism of RigLSTM can greatly help the agent identify its current state and move to the target faster. Thus, RigLSTM often outperforms RIM in this environment.

%\subsection{Further Discussion on Experiments}
%As shown in all the four experiments, our approach consistently outperforms the state-of-the-art RIM and GridLSTM across all the four diversified tasks with a generalization test-bed, especially on the sequential MNIST and CIFAR-10 classification tasks as well as digit copying task, given the fact that classification and memorization can be normally more deterministic than temporal prediction and gaming.

\subsection{Experiments on Multi-digit Copying Task}
%\vspace{-1mm}
Digits memorization tasks are widely utilized in evaluating the abilities of recurrent neural networks. Typical memorization tasks, such as the copying task, usually require recurrent models to reproduce the given sequence after receiving a long sequence of blanks to evaluate if a model is able to memorize the input digit sequence.
% In this tasks, recurrent models to memorize a short sequence of digits,
%  Roughly, this task can be viewed as consisting two independent functions: one to memorize input digits and the other to process the dormant part. Thus, this task is regarded as a typical task to test modularization.
We aim at evaluating the generalization ability when existing environment changes and robustness against complex inputs. And we design a new task for such a purpose.
\begin{figure}[tb!]
\centering
\includegraphics[width=0.44\textwidth]{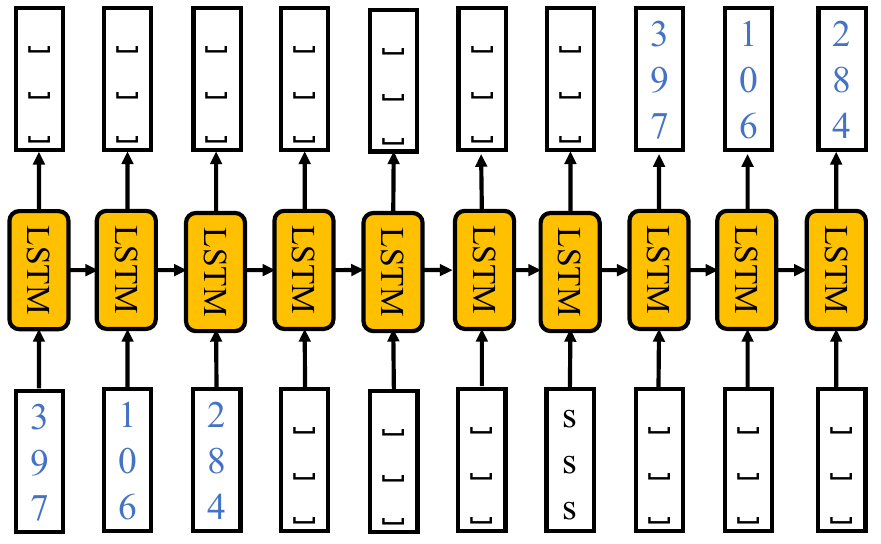}
%\vspace{-10pt}
\caption{
A simple example of 3-digit Copying Task, with $L=3$, $P=3$. At every time step before the dormant phase, 3 digits are fed into the recurrent models.  Once the model receive the special token, it starts to produce the output sequence.
The accuracy is computed based on the generated sequence.}
\label{fig:copying}
%\vspace{-2mm}
\end{figure}

\subsubsection{Task Description}

Our task is to memorize a multi-digit sequence. In this new task, the input at each time step is no longer a single digit, but $M$ digits. The model receives a sequence of multi-digit followed by a sequence of blanks, called \textit{dormant phase}. Then, the model needs to reproduce the input sequence. The goal of models is to reproduce the input sequence. 
Formally, the input sequence is denoted as $A=\{\bm a_1,\bm a_2,\cdots,\bm a_L\}$, where $\bm a_i$ is the input at the $i$-th time step and $\bm a_i$ is a sequence of $M$ digits. The dormant phase is a sequence of blanks of length $P$. A special token is fed into the model after the $M$-digit sequence and dormant phase as a sign for model to produce the output sequence. An example is illustrated in Fig.~\ref{fig:copying}.

% which we call a token.  In our copying task, the input and output sequences of the recurrent model are both in length of $2L+P$, where the first $N$ tokens of the input sequence are target sequence $A=\{\bm a_1,\bm a_2,\cdots,\bm a_N\}$, and the next $P-1$ tokens are blanks, \textit{i.e.}, \textit{dormant phase}. 
% The $P$-th item of the input is a special token \emph{s}. The model is asked to reproduce $A$ immediately after receiving \emph{s}, that is, the last $N$ tokens of the output sequence are supposed to be the same as $A$. All other positions in input and output sequences are filled with blanks. 
% For example, a typical input-output sequence pair would be $\{\bm a_1,\bm a_2,\cdots, \bm a_N, 
% \Vtextvisiblespace[0.8em], 
% \cdots, 
% \Vtextvisiblespace[0.8em], s,
% \Vtextvisiblespace[0.8em], \cdots, \Vtextvisiblespace[0.8em] \}$ 
% and $\{\Vtextvisiblespace[0.8em], \cdots, \Vtextvisiblespace[0.8em], \bm b_1, \bm b_2,\cdots, \bm b_N \}$, 
% where $\bm b_i$ is supposed to reproduce $\bm a_i$, $s$ is a special symbol indicating the copying start, and \Vtextvisiblespace[0.8em] is the blank. 

\begin{table*}[!t]
\centering
\caption{Testing accuracy ($\%$) on Multi-digit Copying Task. Models are trained with dormant length 50 and evaluated in three dormant length settings ($P=100, 200, 400$) respectively.
}
%\vspace{-10pt}
%\small
\begin{tabular}{l|c|c|c|c|c}
		\toprule
% 		Models & ($N, K_a$) &  \tabincell{c}{Token\\ Length} &  \tabincell{c}{Test (100)\\ Acc} &  \tabincell{c}{Test (200)\\ Acc} &  \tabincell{c}{Test (400)\\ Acc} \\
		Models & ($N, K_a$) &  {$M$} &  {Test Acc ($P=100$)} &  {Test Acc  ($P=200$)} &  {Test Acc  ($P=400$)} \\
		\midrule
		LSTM~\cite{lstm} & - & 1 & 45.38 & 25.19 & 16.09 \\
	{GridLSTM}~\cite{gridlstm} & - & 1 & 50.26 & 28.43 & 20.36 \\
		RIM~\cite{goyal2019recurrent} & (6, 4) & 1 & 97.75 & 95.14 & 84.85 \\
		RigLSTM & (6, 4) & 1 & \textbf{99.06} & \textbf{98.89} & \textbf{97.66} \\
		\midrule
		LSTM~\cite{lstm} & - & 2 & 29.74 & 11.51 & 10.70 \\
	{GridLSTM}~\cite{gridlstm} & - & 2 & 31.25 & 15.37 & 14.26 \\
		RIM~\cite{goyal2019recurrent} & (6, 4) & 2 & 94.45 & 89.30 & 79.59 \\ 
		RigLSTM & (6, 4) & 2 & \textbf{98.28} & \textbf{95.43} & \textbf{92.63} \\
		\midrule
		LSTM~\cite{lstm} & - & 4 & 25.07 & 12.81 & 10.06 \\
	{GridLSTM}~\cite{gridlstm} & - & 4 & 29.33 & 15.94 & 12.11 \\
		RIM & (6, 4) & 4 & 75.91 & 70.80 & 63.23 \\ 
		RigLSTM & (6, 4) & 4 & \textbf{95.61} & \textbf{92.46} & \textbf{89.95} \\
		\midrule
		LSTM~\cite{lstm} & - & 8 & 18.52 & 11.74 & 10.07 \\
	{GridLSTM~\cite{gridlstm}} & - & 8 & 21.25 & 14.09 & 12.75 \\
		RIM~\cite{goyal2019recurrent} & (6, 4) & 8 & 65.46 & 52.26 & 42.27 \\ 
		RigLSTM & (6, 4) & 8 & \textbf{90.24} & \textbf{87.95} & \textbf{84.59} \\
% 		\specialrule{0em}{1pt}{1pt}
% 		\hline
% 		\specialrule{0em}{1pt}{1pt}
% 		\hline
    \bottomrule
	\end{tabular}
\normalsize
\label{tab:copying}
%\vspace{-2mm}
\end{table*}

\subsubsection{Experimental Settings}
For generalization ability testing, the recurrent models are trained with dormant phase length setting to 50, and evaluated  with dormant phase lengths equal to 100, 200 and 400. We set $M$ with varying values, \textit{i.e}, $1, 2, 4$ and $8$, to evaluate their abilities when the task becomes difficult. The sequence length $L$ is fixed to 10, and all digits are randomly sampled from $[0,9]$. We take LSTM and RIM as our baseline models, and compare them with our model in terms of testing accuracy.

\subsubsection{Analysis and Discussion}
The results are shown in Table~\ref{tab:copying}. We can see that our model achieves the best performance under all the settings. For most cases, the performances of LSTM and GridLSTM are rather poor, which proves that special designs are necessary for the testing cases when environment changes. When $M$ is small, \textit{e.g.}, 1 and 2, the task is relatively easy. Both RIM and RigLSTM provide satisfying generalization abilities. However, when the task  becomes harder, our model is far better than RIM and the gap also become larger. Thus, RigLSTM performs significantly better than RIM for complex inputs. This might be attributed to the input selection and hidden state design, which makes the cells in our model better at  selecting  relevant information as inputs. Moreover, as the length of dormant length becomes larger, the performance of all models drops. But RigLSTM can still achieve accuracy $84\%$ for the test with dormant phase length set to 400 and $M=8$, while the accuracy of RIM is only $42\%$. It suggests that RigLSTM is more robust to the changes of dormant phase length.  This superiority of our model in the generalization ability may be due to the soft state updating mechanism. Compared with RIM, our state updating design is more effective for propagation of information from previous states. Overall, the designs of cell selection, input transformation and feature selection, hidden state selection and soft state updating greatly improve the generalization ability when  environment changes, and help our model achieve far better performance than RIM.

\section{Conclusion}\label{sec:con}
We have proposed the recurrent independent Grid LSTM (RigLSTM) for generalizable sequence learning. It aims to decouple and streamline the modeling of underlying modular subsystems with sparse interaction over time. 
RigLSTM is composed of a group of independent LSTM cells that can actively cooperate with each other. We proposed
cell selection, input selection, hidden state selection, and soft state updating, integrated with the Grid LSTM, to exploit the underlying modular structures on sequence modeling tasks. The cell selection forces the cells inside the proposed unit to form specialisms. 
The input selection helps the LSTM cells to select only relevant input information, the hidden state selection facilitate the LSTM cells to absorb only relevant context information. By virtue of the proposed components, our model enjoys competitive generalization and robustness, which has been validated on diversified sequence modeling tasks. The source code will be made publicly available.

% \textbf{Limitations.} Our method is based on the assumption that the dynamic system can be decomposed into a few subsystems, which may not work well when this condition is broken.
% \label{limitations}

% \textbf{Potential Negative Societal Impacts.} The proposed sequential model can be applied in some areas like personalized recommendation where the privacy can be a concern.
% \label{negative}

% or


\appendices

\ifCLASSOPTIONcompsoc
  \section*{Acknowledgments}
\else
  \section*{Acknowledgment}
\fi
This work was partly supported by National Key Research and Development Program of China (2020AAA0107600), National Natural Science Foundation of China (61972250, 72061127003), and Shanghai Municipal Science and Technology (Major) Project (22511105100, 2021SHZDZX0102).

\ifCLASSOPTIONcaptionsoff
  \newpage
\fi



%\bibliographystyle{IEEEtran}

\bibliographystyle{IEEEtran}
\bibliography{ref}

\end{document}